\documentclass{article}
\usepackage{ijcai25}

\usepackage{times}
\usepackage{soul}
\usepackage{url}
\usepackage[hidelinks]{hyperref}
\usepackage[utf8]{inputenc}
\usepackage[small, labelsep=period]{caption}
\usepackage{graphicx}
\usepackage{amsmath}
\usepackage{amsthm}
\usepackage{booktabs}
\usepackage{algorithm}
\usepackage{algorithmic}
\usepackage[switch]{lineno}

\usepackage{xcolor}
\usepackage{amsmath} 
\usepackage{amssymb}
\usepackage{multirow}
\usepackage{caption}
\usepackage{subcaption}
\usepackage{amsfonts}
\usepackage[parfill]{parskip}

\usepackage{amsmath}
\usepackage{amssymb}
\usepackage{mathtools}
\usepackage{amsthm}

\usepackage[capitalize,noabbrev]{cleveref}
\usepackage{algorithm}
\usepackage{algorithmic}

\theoremstyle{plain}

\theoremstyle{definition}

\theoremstyle{remark}

\newtheoremstyle{theorstyle}
{.1em} % Space above
{.1em} % Space below
{} % Body font
{} % Indent amount
{\bfseries} % Theorem head font
{.} % Punctuation after theorem head
{.5em} % Space after theorem head
{} % Theorem head spec (can be left empty, meaning `normal')

\theoremstyle{theorstyle} 
\theoremstyle{theorstyle} \newtheorem{objective}{Objective}[section]

\usepackage{amsmath,amsfonts,bm}

\def\eqref#1{equation~\ref{#1}}
\def\1{\bm{1}}

\DeclareMathAlphabet{\mathsfit}{\encodingdefault}{\sfdefault}{m}{sl}
\SetMathAlphabet{\mathsfit}{bold}{\encodingdefault}{\sfdefault}{bx}{n}

\title{\textbf{Robult}: Leveraging Redundancy and Modality-Specific Features for Robust Multimodal Learning}

\author{Duy A. Nguyen$^{1, 3}$ \and
Abhi Kamboj$^{2}$\and
Minh N. Do$^{2, 3}$ \\
\affiliations
$^1$Siebel School of Computing and Data Science, UIUC, US \\
$^2$Department of Electrical and Computer Engineering, UIUC, US\\
$^3$VinUni-Illinois Smart Health Center, VinUniversity, Vietnam
\emails
\{duyan2, akamboj2, minhdo\}@illinois.edu
}

\date{Jan 2025}
\begin{document}

\maketitle

\begin{abstract}
Addressing missing modalities and limited labeled data is crucial for advancing robust multimodal learning. We propose \textbf{Robult}, a scalable framework designed to mitigate these challenges by preserving modality-specific information and leveraging redundancy through a novel information-theoretic approach. Robult optimizes two core objectives: (1) a soft Positive-Unlabeled (PU) contrastive loss that maximizes task-relevant feature alignment while effectively utilizing limited labeled data in semi-supervised settings, and (2) a latent reconstruction loss that ensures unique modality-specific information is retained. These strategies, embedded within a modular design, enhance performance across various downstream tasks and ensure resilience to incomplete modalities during inference. Experimental results across diverse datasets validate that Robult achieves superior performance over existing approaches in both semi-supervised learning and missing modality contexts. Furthermore, its lightweight design promotes scalability and seamless integration with existing architectures, making it suitable for real-world multimodal applications.
\end{abstract}

\section{Introduction}
\label{sec:intro}

\textbf{Motivation:} 
In the Big Data era, multimodal learning significantly improves data exploitation, outperforming single-modality approaches \cite{mm_sup_um}. However, most existing methods \cite{interaction_1,interaction_2} operate under idealized assumptions: fully labeled training datasets and consistently available modalities during evaluation. In practice, the challenges of missing modalities and semi-supervised learning often coexist, yet current research typically addresses these problems in isolation. For example, missing modalities may arise when autonomous vehicles lose sensor inputs due to environmental obstructions or when medical diagnostics in resource-limited settings lack access to all imaging modalities. Simultaneously, the scarcity of labeled data across domains remains a critical bottleneck, particularly in multimodal contexts where individual modalities often require specialized labeling.

Addressing these challenges independently limits the adaptability of multimodal systems and fails to capture the complexities of real-world deployments. This work uniquely addresses these dual challenges by integrating strategies to handle incomplete modalities and leverage unlabeled data simultaneously. By doing so, our approach enhances both flexibility and applicability, enabling robust performance in diverse and imperfect scenarios. Unlike current literature, which rarely addresses both issues concurrently, this work bridges a critical gap with an innovative and unified solution.

\textbf{Existing literature:} 
One of the primary challenges in multimodal learning is handling corrupted or missing modalities. 
% For instance, an autonomous vehicle may lose access to camera data obscured by mud, relying solely on lidar, or a medical diagnostic system in resource-constrained environments may depend on a single imaging modality. 
Existing methods to address missing modalities typically fall into two categories:
\begin{itemize}
    \item[1.] \textbf{Generative approaches}, such as VAE-based models \cite{vae_1}, which reconstruct missing modalities. While recent advancements \cite{nips22,aaai23} demonstrate promising performance, these methods often depend on specific architectures, limiting their flexibility. 
    \item[2.] \textbf{Transfer learning methods}, which align latent spaces for cross-modal knowledge transfer \cite{cvpr22,missing_1,kdd20}. These methods focus on adaptable training strategies \cite{cvpr23} but often lack a strong theoretical foundation and are primarily guided by empirical intuition.
\end{itemize}

Simultaneously, the need for semi-supervised learning arises from practical challenges in labeling raw data, especially in domains where annotations are scarce or labor-intensive to obtain. This challenge is amplified in multimodal learning, as each modality may require distinct expertise for labeling. For example, tasks such as object segmentation across video and lidar data in autonomous driving \cite{autonomous_1} or medical segmentation across imaging modalities \cite{healthcare_3} demand diverse and often non-standardized labeling procedures. Recent advancements in semi-supervised learning, such as knowledge distillation \cite{semi_6} and pseudo-labeling \cite{semi_1}, have shown promise, but these techniques are often designed for specific applications and struggle to generalize across varied multimodal settings.

A detailed discussion of existing literature in both areas is provided in Appendix \ref{sec:literature}.

\textbf{Proposed approach:} 
\begin{figure}
    \centering
    \includegraphics[width=.8\columnwidth]{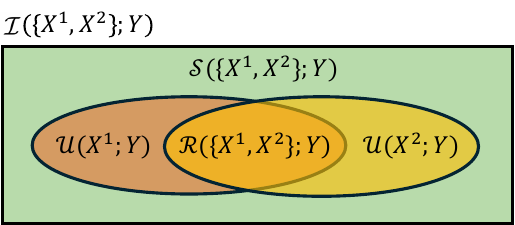}
    \caption{Partial Information Decomposition for 2 input modalities with target variable.}
    \label{fig:pid}
\end{figure}
The dual challenges of missing modalities and semi-supervised learning remain critical open problems in multimodal research. This work advances transfer learning methods for missing modalities (category 2) that pertain architectural flexibility and dataset compatibility, while introducing a novel design to reduce dependence on labeled data, ensuring robustness and adaptability in real-world scenarios.

Using Partial Information Decomposition \cite{pid}, the mutual information provided by an input $X$ with $M$ modalities ($X^1, \ldots, X^M$) for a given task $Y$ can be decomposed into:
\begin{equation}
\begin{aligned}
    \mathcal{I}(&\left\{X^1, \ldots, X^M\right\}; Y) = \mathcal{R}(\left\{X^1, \ldots, X^M\right\}; Y) \\
    &+ \sum_{i=1}^M \mathcal{U}(X^i; Y) + \mathcal{S}(\left\{X^1, \ldots, X^M\right\}; Y),
\end{aligned}
\end{equation}
where $\mathcal{R}$ represents redundancy (shared task-relevant information among $M$ modalities), $\mathcal{U}$ denotes unique information specific to the $i^\text{th}$ modality, and $\mathcal{S}$ quantifies synergy, the additional knowledge generated through interactions among modalities (Figure \ref{fig:pid} illustrates PID with 2 modalities).

In an ideal setting with access to all modalities, a fusion technique would efficiently capture $\mathcal{R}$ and $\mathcal{S}$ to optimize predictions for $Y$. However, in real-world scenarios where some modalities are unavailable, replicating $\mathcal{S}$ becomes challenging. For example, with two modalities, $Y = \texttt{linear}(X^1, X^2) = \texttt{non\_linear}(X^1)$ demonstrates that synergy ($\mathcal{S}$) improves predictions when both $X^1$ and $X^2$ are available. Access to $X^1$ alone makes the relationship more complex and harder to model with deep networks.

Approaches like knowledge distillation \cite{cvpr23} and contrastive learning \cite{clip} aim to address missing modalities by mimicking the representations produced by fused modalities in their absence. From an information-theoretic perspective, these methods focus on replicating redundant information ($\mathcal{R}$) through latent-space alignment.
Building on this foundation, we explicitly introduce a mutual information maximization objective (Objective \ref{obj:mi}) to align unimodal and fused representations. This alignment ensures efficient knowledge transfer while minimizing reliance on labeled data using a novel soft-positive pseudo-labeling mechanism that accounts for pseudo-label uncertainty.

While alignment effectively captures $\mathcal{R}$, it can unintentionally diminish unique information ($\mathcal{U}$) from individual modalities. This loss is particularly detrimental in semi-supervised settings or when other modalities are unavailable. To address this, we propose preserving modality-specific information ($\mathcal{U}^i$) via Objective \ref{obj:entropy}. This is achieved using a simple yet effective reconstruction procedure in the latent space, universally applicable across modalities. Our results and ablations (Tables \ref{tab:mosi_mosei} and \ref{tab:ablation}) demonstrate how retaining $\mathcal{U}^i$ improves performance.

Together, Objectives \ref{obj:mi} and \ref{obj:entropy} form the foundation of our semi-supervised multimodal learning method, \textbf{Rob}ust M\textbf{ult}imodal Pipeline (\textbf{Robult}). Robult effectively balances redundancy alignment and unique information preservation, ensuring accuracy and robustness in scenarios with missing modalities.
Empirical results (Section \ref{sec:experiment}) and theoretical underpinnings highlight Robult's superior performance compared to existing methods, demonstrating its adaptability to diverse real-world settings.

\textbf{Contributions.} Our primary contributions are:
\begin{itemize} 
    \item Jointly addressing the dual challenges of missing modalities and semi-supervised learning.
    \item Framing two objectives under an information-theoretic perspective and deriving novel loss functions to achieve these goals.
    \item Introducing a soft Positive-Unlabeled contrastive loss that efficiently utilizes limited labeled data through selective weighting of potential positives.
\end{itemize}

\section{Methodology}
\label{sec:method}
\begin{figure}[!t]
    \centering
    \includegraphics[width=\linewidth]{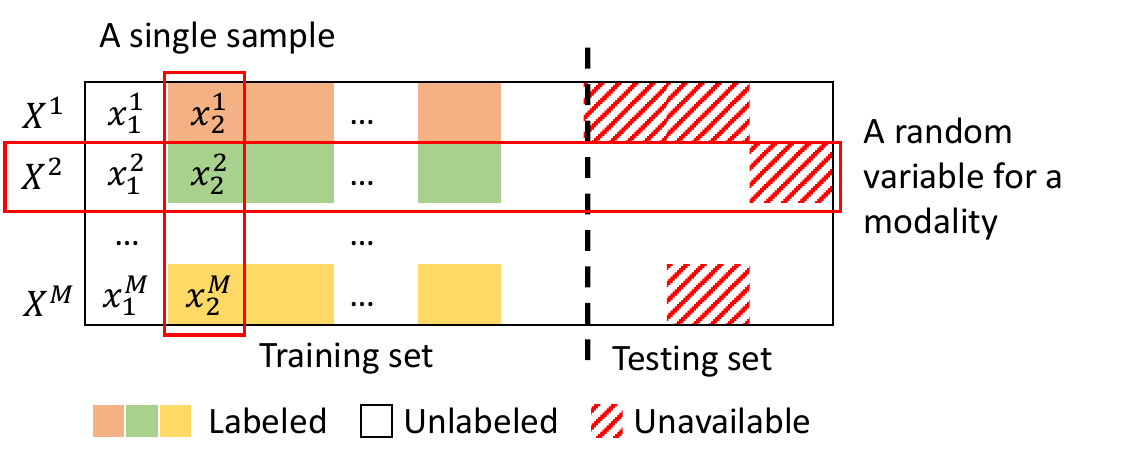}
    \caption{Training/evaluation datasets under investigation.}
    \label{fig:training_setting}
\end{figure}

\begin{figure}[!t]
    % \begin{minipage}{0.55\linewidth}
    \centering
    \begin{subfigure}[b]{0.5\textwidth}
    \centering
    \includegraphics[width=\textwidth]{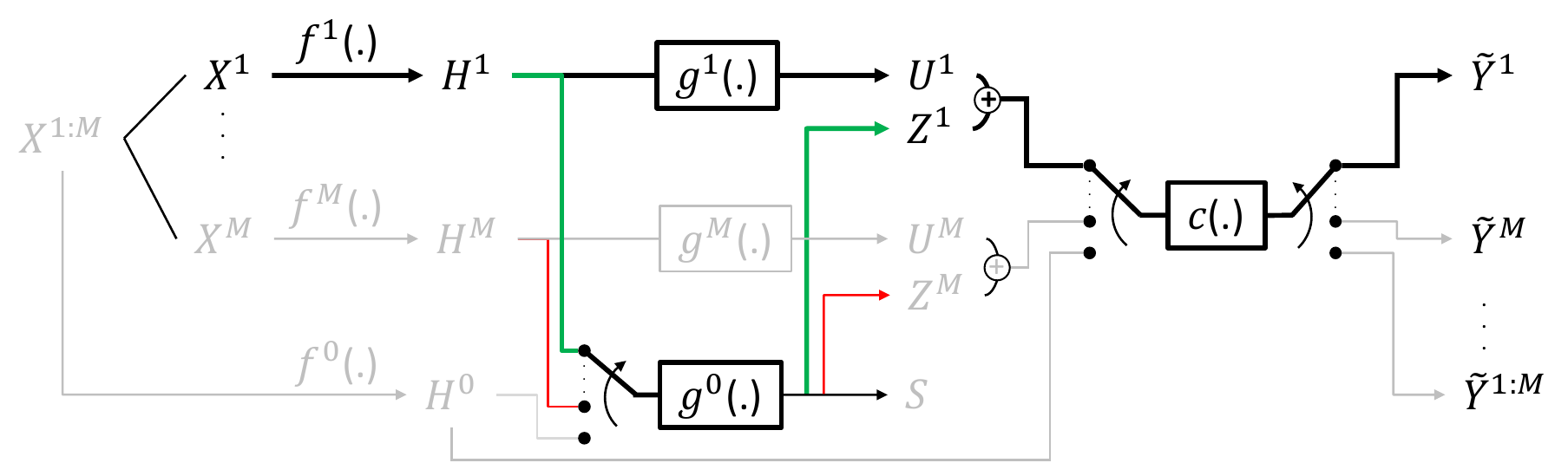}
    \caption{Inference when some modalities are missing}
    \label{fig:overall_some}
    \end{subfigure}
    \hfill
    \begin{subfigure}[b]{0.49\textwidth}
    \centering
    \includegraphics[width=\textwidth]{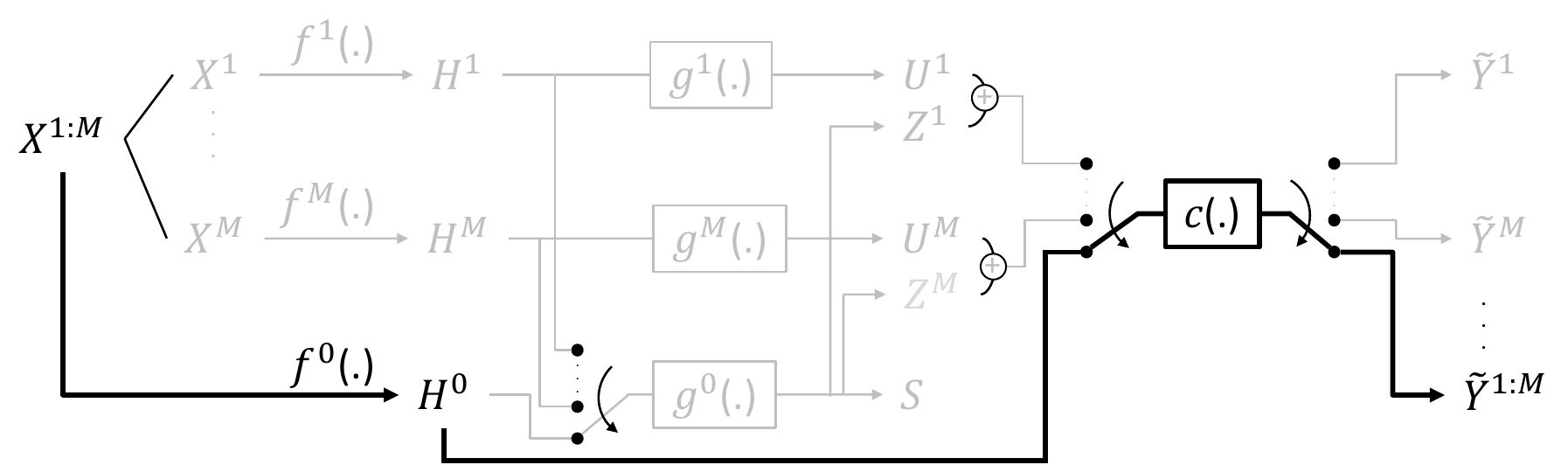}
    \caption{Inference when all modalities are available}
    \label{fig:overall_full}
    \end{subfigure}
    \caption{
    (\ref{fig:overall_some}) For each available modality $i \in \left\{ 1,\dots, M\right\}$, $f^i(.)$ extracts the latent representation $h_j^i$. The modules $g^i(.)$ and $g^0(.)$ then extract the unique information $u_j^i$ and redundant representation $z_j^i$, respectively. These are combined and processed by the shared module $c(.)$ to predict $\tilde{y}_j^i$. Late fusion strategies aggregate these outputs into the final prediction $\tilde{y}_j$.
    (\ref{fig:overall_full}) When all modalities $x_j^{1:M}$ are available, $f^0(.)$ extract a fused latent vector $h^0_j$. This vector is directly passed through $c(.)$ to generate the final output $\tilde{y}_j^{1:M}$.
    }
    \label{fig:overall}
\end{figure}
\label{sec:problem}
\textbf{Training setting:} We consider a training scenario where each sample includes all modalities, but only some samples have corresponding labels (Figure \ref{fig:training_setting}). 
Formally, let the training dataset be $\mathcal{D}_{train} = \left\{(x_1, y_1), \ldots, (x_k, y_k), x_{k+1}, \ldots x_n\right\}$, where each data point $x_j = \left(x^1_j, \dots, x^M_j \right)$ contains $M$ modalities. The dataset consists of $n$ samples, of which $k$ are labeled ($0 \le k < n$).

\textbf{Evaluation setting:} 
To mimic real-world deployment, the evaluation dataset $\mathcal{X}_{test} = \left\{ x_1, \ldots, x_m \right\}$ consists of samples with potential missing modalities - e.g. $x_j = (x_j^a \text{'s}); \forall a \in \mathcal{A}^M_j \subseteq \left\{1, \ldots, M \right\}$. Here, $\mathcal{A}^M_j$ denote the indices of all available modalities for sample $x_j$ (Figure \ref{fig:training_setting}).

\textbf{Notation:} For clarity, in the theoretical framework, we denote the input random variables corresponding to the $i^{th}$ modality as $X^i$ (Figure \ref{fig:training_setting}). We would refer back to the observation-level notation $x^i_j$ whenever needed. 

Figure \ref{fig:overall} illustrates our versatile multimodal pipeline named Robult. Robult consists of $M$ modality-specific branches and a fusion branch indexed with zero. 
For each input variable $X^i$, it is first projected into a shared latent space:
$$
    H^i = f^i(X^i), \quad i \in \left\{1, \dots, M \right\}.
$$
In parallel, a fusion network $f^0(.)$ processes all input modalities jointly to produce a fused latent variable (also reside on the shared latent space):
$$
    H^0 = f^0(X^{1:M}),
$$
where index 0 denotes the joint latent variable using all modalities. This latent-space projection provides two key benefits:
\begin{itemize}
    \item \textbf{Generalization}: Robult supports different combination of modalities and their preferred projection methods/architectures.
    \item \textbf{Efficient Fusion}: Various strategies can be adopted before Robult's main logic to efficiently attain fused representations.
\end{itemize}
Each modality-specific module $g^i(.)$ extracts unique information of corresponding modality:
$$
U^i = g^i(H^i), \quad i \in \left\{1, \ldots, M \right\},
$$
while the shared module $g^0(.)$ effectively captures the redundant information by processing individual or joint latent variables:
$$
Z^i = g^0(H^i), S = g^0(H^0). 
$$

\textbf{Objectives:} 
% We consider the setting where labels are scarce during training in addition to potentially missing modalities during testing. 
As discussed in Section \ref{sec:intro}, we draw inspiration from the second category of literature on missing modalities, which primarily leverages knowledge distillation and contrastive loss techniques \cite{icml22}. These methods aim to align unimodal (student) representations, such as $Z^i$, with the fused (teacher) representation $S$, enabling unimodal representations to efficiently replicate the redundant information encapsulated in the fused representation. Although this redundant information exists within each modality, extracting it directly from a single modality can be significantly more complex without access to all modalities. For example, encoding object shape from an image may be challenging, whereas a textual description can explicitly provide the same information. Mimicking the fused representation offers a shortcut, simplifying the extraction of redundant information within unimodal representations.

However, these approaches fail in the semi-supervised training setting, owing to 2 emerging challenges: 
\begin{itemize}
    \item (1) The \emph{over reliance on label signals} (e.g. classification clusters) in the alignment process.
    \item (2) the \emph{diminishing of unique information} contained in different modality after alignment process. 
\end{itemize}
Regarding the former challenge, we directly address the label need of Contrastive Learning with Label-level sampling \cite{s_cl_2} by a novel soft Positive-Unlabelled (PU) contrastive loss, together with an adaptive weighting strategy, detailed about which is covered in Section \ref{sec:soft_pu}. 
Theoretically, this PU contrastive loss corresponds to a mutual information maximization problem between the desired fused representation $S$ and the learned unimodal representation projected into that same latent space $Z^i$'s. Objective \ref{obj:mi} of our method can be expressed as follow:

\begin{objective}
    \label{obj:mi}
    Aligning $S$ and $Z^i$ by maximizing the mutual information $\mathcal{I}(S, Z^i)\quad (i= \overline{1, M})$.
\end{objective}

For the latter challenge, we observe that attempting to only align modalities during training diminishes the unique information provided by each modality, thus the model is losing information that could help inform it when only that specific modality is available. Therefore, a model should ideally benefit by maintaining the redundancy while also explicitly preserving each modality's unique information.
This claim is later supported by experimental results and ablations (Section \ref{sec:experiment} - Table \ref{tab:mosi_mosei}). 
To address the challenge of vanishing modality-specific information from multimodal alignment during training, we emphasize a disentanglement strategy that preserves unique information while still facilitating the redundancy learning process of Objective \ref{obj:mi}. 
Robult integrates a set of modules $g^i(H^i)$, where $i=1, \dots, M$, to produce unique representations $U^i$ for each modality. 
We aim to preserve the unique information for each modality via the learning of $U^i$ with Objective \ref{obj:entropy} as follows:

\begin{objective}
\label{obj:entropy}
Learning $U^i$ by minimizing the conditional entropy $\mathcal{H}(H^i | Z^i, U^i)\quad (i= \overline{1, M})$.
\end{objective}

During training, all branches are executed and three loss functions update the learned modules. The soft positive unlabeled loss, $\mathcal{L}_{PU}$, maximizes knowledge extraction from the few labeled samples in a batch (Subsection \ref{sec:soft_pu}).
The reconstruction loss, $\mathcal{L}_{rec}$ forces each branch to extract unique modality-specific information (Subsection \ref{sec:reconstruction}).
The task-specific supervised loss, $\mathcal{L}_{sup}$ is used on the labeled samples across all modules to learn label information (Subsection \ref{sec:training_strat}). 

\subsection{Maximizing Mutual Information with Soft Positive-Unlabeled Contrastive Learning}
\label{sec:soft_pu}
To address the impact of missing modalities, we aim to learn redundant information by aligning the fused latent variable $S$ with unimodal representations $Z^i$, as formalized in Objective \ref{obj:mi}. This is achieved by maximizing their mutual information $\mathcal{I}(S, Z^i)$. However, direct computation of this quantity is infeasible without access to the joint distribution $p_{S, Z^i}$ or the marginal distributions $p_S$ and $p_{Z^i}$. Instead, we derive and optimize a lower bound for this mutual information.

\textbf{Lower Bound Derivation.} Let $F$ be a binary random variable indicating whether a pair $(s_j, z^i_k)$ 
% ($j$ could be different from $k$ - meaning the representation pair is generated from two independent samples)
is sampled from the joint distribution $p_{S, Z^i}$ ($F=1$) or from the product of marginal distributions $p_S \otimes p_{Z^i}$ ($F=0$). Then, a lower bound for $\mathcal{I}(S, Z^i)$  is expressed as:
\begin{equation}
\begin{aligned}
    \mathcal{I}(S, Z^i) &\ge -\mathbb{E}_{p_{S, Z^i}}\log v(S,Z^i) \\
    &= -\mathbb{E}_{p(S, Z^i|F=1)}\log v(S,Z^i)
\end{aligned}
\label{eq:res_1}
\end{equation}
where $v(S,Z^i)$ is a non-parametric approximation of $p(F=1| S,Z^i)$. For a sampled pair $(s_j, z^i_k)$ in a batch of $B$ samples, where $s_j$ is the fused representation for sample $j$ and $z^i_k$ is the $i^{th}$ modality-specific representation for sample $k$, $v(s_j,z_k^i)$ is defined as:
$$
\begin{aligned}
    v(s_j,z^i_k) = \frac{\phi(s_j, z_k^i)}{\sum_h^B\phi(s_j, z_h^i)}; \quad \\
    \text{where}\quad \phi(s_j, z_k^i) = exp(\langle s_j; z_k^i \rangle / \tau).
\end{aligned}
$$
The derivation of Result \ref{eq:res_1} is detailed in Appendix \ref{sec:sup_mi}.

\textbf{Challenges with Sampling.} The lower bound in \eqref{eq:res_1} relies on expectations under $\mathbb{E}{p_{S, Z^i}}$, a key source of deviation in existing studies, which presents challenges in practice. Two common sampling strategies include:
\begin{itemize}
    \item[1.] \textbf{Instance-level sampling}, which considers only intra-sample pairs $(s_j, z^i_j)$ within q mini-batch \cite{clip}.
    \item[2.] \textbf{Label-level sampling}, which uses label information to sample inter-sample pairs $(s_j, z^i_k)$ with the same labels, e.g. $y_j = y_k$ \cite{s_cl_2}.
\end{itemize}
The first approach risks introducing false negatives, while the second requires fully labeled data, limiting its applicability in semi-supervised settings.

\textbf{Soft Positive-Unlabeled (PU) Contrastive Loss.} To overcome these limitations, we propose a novel soft Positive-Unlabeled (PU) contrastive loss with adaptive weighting. Let $L$ indicate whether a sampled pair $\langle s_j, z^i_k \rangle$ is labeled $(L=1)$ or not $(L=0)$ (where $L=1$ if the label information of both samples $i$ and $k$ is known, and $L=0$ otherwise), The lower bound in \eqref{eq:res_1} can then be decomposed as:

\begin{equation}
\small
\begin{aligned}
-\mathbb{E}_{p_{S, Z^i}} \log v(S, Z^i)
&= -\mathbb{E}_{p(S, Z^i \mid F=1)} \log v(S, Z^i) \\
&= -p(L{=}1)\mathbb{E}_{p(S, Z^i \mid F{=}1, L{=}1)} \log v(S, Z^i) \\
&\quad -p(L{=}0)\mathbb{E}_{p(S, Z^i \mid F{=}1, L{=}0)} \log v(S, Z^i)
\end{aligned}
\label{eq:res_2}
\end{equation}

We formulate the two terms in \eqref{eq:res_2} as separate loss components: $\mathcal{L}_{lb}$ for labeled data and  $\mathcal{L}_{ulb}$ for the unlabeled data.

For labeled samples, let ${B^j_{1,1}}$ denote the index set of inputs in the batch that share the same class as sample $j$ ($F=1$) and are labeled ($L=1$), i.e. $k \in B^j_{1, 1} \iff (s_j,z^i_k) \sim p(S, Z^i | F=1, L=1)$.
The labeled loss is then defined as a NT-Xent-like contrastive loss \cite{simclr}:
\begin{equation}
    \begin{aligned}
        &\mathcal{L}_{lb} = \frac{1}{M} \sum_{i=1}^M \sum_{j \in B} \frac{1}{||B^j_{1, 1}||} \sum_{\substack{{k \in B^j_{1, 1}}}}\log v(s_j,z^i_k).
    \end{aligned}
\end{equation}
\begin{figure}[!t]
    \centering
    \includegraphics[width=0.8\linewidth]{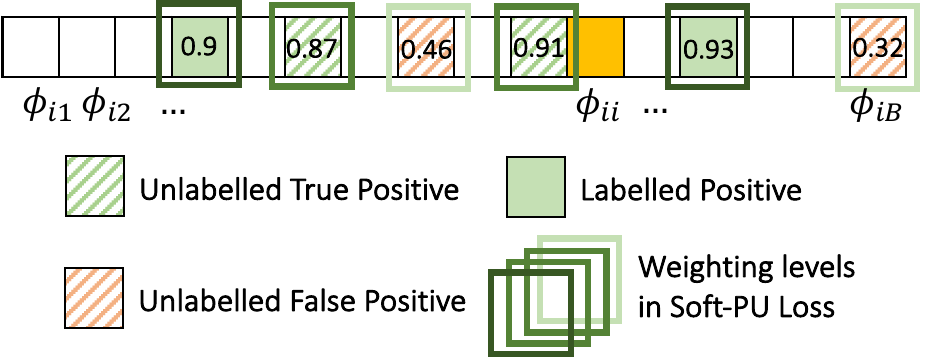}
    \caption{Soft-PU Loss mechanism. Unlabeled positive pairs are identified using soft labels from Robult's classifier. These pairs are re-weighted based on their proximity and the mean proximity of true labeled positive pairs to mitigate false positives.}
    \label{fig:soft_pu}
\end{figure}

For unlabeled samples, directly sampling from $p(S, Z^i | F=1, L=0)$ is challenging due to the absence of label information. To address this, we propose leveraging the output of the Robult classifier to generate soft labels, regularized by adaptive weights calculated dynamically within each mini-batch. This approach effectively balances the influence of soft-labeled pairs and mitigates the limitations of traditional pseudo-labeling methods \cite{semi_1}.

A key challenge during initial training stages is the instability of the Robult classifier, which may produce unreliable outputs and hinder effective filtering of false positives, as illustrated in Figure \ref{fig:soft_pu}. To counteract this, we adjust the contribution of soft-labeled pairs to the loss function, ensuring robust training. 
For each anchor sample $s_j$ within a mini-batch, there are labeled positive partners or, in unsupervised scenarios, unimodal representations $z^i_j$. The average proximity of these labeled partners to $s_j$ serves as a reference for determining ``positive'' proximity thresholds. Unlabeled positive pairs identified by the Robult classifier are expected to exhibit proximity close to this reference mean. To refine the loss computation, we increase the weight of pairs that align closely with the reference mean and decrease the influence of outliers. This weighting mechanism, implemented using an RBF kernel, allows precise adjustment of each couplet's contribution based on its proximity. By dynamically adapting weights, our method effectively reduces the impact of potential false positives and enhances the overall performance of the final loss function, as demonstrated in Figure \ref{fig:soft_pu}.

\begin{equation}
    \begin{aligned}
        w_{jk}^i &= RBF(\phi_{j}^i, \phi(s_j, z^i_k)); \\
        \quad \text{where} \quad\phi_{j}^i &= mean\left\{\phi(s_j, z^i_{\Tilde{k}}) | \Tilde{k} \in B^j_{1, 1} \right\}.
    \end{aligned}
\end{equation}
Let ${B^j_{1, 0}}$ denote the index set of inputs in the batch that share the same class as anchor $j$ ($F=1$) but are not labeled ($L=0$), i.e. $k \in B^j_{1, 0} \iff (s_j,z^i_k) \sim p(S, Z^i | F=1, L=0)$.
The unlabeled loss is given by:
\begin{equation}
    \begin{aligned}
        \mathcal{L}_{ulb} &= \frac{1}{M} \sum_{i=1}^M \sum_{j\in B} \frac{1}{||B^j_{1, 0}||} \sum_{\substack{k \in B^j_{1, 0}}}w_{jk}^i\log v(s_j,z^i_k).
    \end{aligned}
\end{equation}
The complete soft Positive-Unlabeled (PU) loss is the sum of the labeled and unlabeled components:
\begin{equation}
    \begin{aligned}
        \mathcal{L}_{PU} =  \mathcal{L}_{ulb} + \mathcal{L}_{lb}
    \end{aligned}
\end{equation}

\subsection{Minimizing Conditional Entropy with Latent Reconstruction Error}
\label{sec:reconstruction}
This section details the procedure for achieving Objective \ref{obj:entropy}, which focuses on preserving unique information $U^i$. Let $p_{U^i, Z^i}$ denote the joint distribution of $U^i$ and $Z^i$, where $(u^i_j, z^i_j) \sim p_{U^i, Z^i}$ are derived from the corresponding instance $h^i_j$. Inspired by \cite{infogan}, the conditional entropy $\mathcal{H}(H^i|Z^i, U^i)$ is expressed as:
\begin{equation}
\small
\begin{aligned}
\mathcal{H}(H^i|U^i, Z^i) = -&\mathbb{E}_{p_{U^i, Z^i}}\Bigl[\mathbb{E}_{p_{H^i \mid U^i, Z^i}}\bigl[\log p\left(H^i \mid U^i, Z^i\right)\bigr]\Bigr]
\end{aligned}
\label{eq:entropy_1}
\end{equation}
\textbf{Upper Bound Derivation.} Since directly computing $p(H^i \mid U^i, Z^i)$ is challenging, we approximate it using a distribution $q(H^i \mid U^i, Z^i)$. Substituting $q$ into Eq. \ref{eq:entropy_1}, we derive:
\begin{equation}
\scriptsize
\begin{aligned}
\mathcal{H}(H^i|Z^i, U^i)
&= -\mathbb{E}_{U^i, Z^i}\Big[\mathbb{E}_{H^i | U^i, Z^i}
\big[\log q(H^i | U^i, Z^i) \cdot \tfrac{p(H^i | U^i, Z^i)}{q(H^i | U^i, Z^i)}\big]\Big] \\
&= -\mathbb{E}_{U^i, Z^i}\Big[\mathbb{E}_{H^i | U^i, Z^i}
[\log q(H^i | U^i, Z^i)] + \mathrm{KL}(p||q)\Big] \\
&\le -\mathbb{E}_{U^i, Z^i}\Big[\mathbb{E}_{H^i | U^i, Z^i}
[\log q(H^i | U^i, Z^i)]\Big]
\end{aligned}
\label{eq:entropy_2}
\end{equation}

The last inequality arises because the KL divergence $\mathrm{KL}(p || q)$ is non-negative. Thus, minimizing $\mathcal{H}(H^i|Z^i, U^i)$ reduces to minimizing its Evidence Lower Bound (ELBO)-like  \cite{vae} upper bound using the approximating distribution $q\left(H^i \mid U^i, Z^i\right)$. 

\textbf{Modeling $q\left(H^i \mid U^i, Z^i\right)$ with Latent Reconstruction.} We model $q\left(H^i \mid U^i, Z^i\right)$ through a latent reconstruction procedure in the shared latent space. Specifically, we define a reconstruction module $r^i(U^i, Z^i) = \Tilde{H}^i$ where $\Tilde{H}^i$ approximates $H^i$. For each pair $(u^i_j, z^i_j) \sim p_{U^i, Z^i}$ generated from $h^i_j$, the module $r^i(.)$ attempts to reconstruct $\Tilde{h}^i_j$ such that it closely resembles $h^i_j$. The reconstruction loss is formulated as:
\begin{equation}
        \mathcal{L}_{rec} = \frac{1}{MB}\sum_{i=1}^M\sum_{j=1}^B 1 - \langle \Tilde{h}^i_j, h^i_j \rangle ^2,
\end{equation}
where $\langle.;.\rangle$ denotes the $L2$-normalized dot product operation, $B$ is the size of mini-batch, and $M$ is the number of modalities. 

This reconstruction in latent space is computationally efficient and alleviates the complexity of directly reconstructing raw modality data. Moreover, it generalizes well across various modalities, enhancing the flexibility of Robult.
The reconstruction loss $\mathcal{L}_{rec}$ is back-propagated exclusively through the $M$ unimodal branches. This design ensures that the unique information $\mathcal{U}(X^i; Y)$ is preserved through $U^i$, without interfering with the shared branch's focus on learning redundant information, as discussed in Section \ref{sec:soft_pu}.

\subsection{Training strategy}
\label{sec:training_strat}
The objectives outlined in Sections \ref{sec:soft_pu} and \ref{sec:reconstruction} are optimized through their respective loss functions. Due to the distinct purposes of these objectives, it is advantageous to learn them separately. Specifically, we apply $\mathcal{L}_{rec}$ to guide the learning of $g^i(.)$ ($i=1,\dots,M$), while $\mathcal{L}_{PU}$ drives the optimization of $f^i(.), f^{0}(.)$, and $g^{0}(.)$. 
To maximize the utility of labeled data, we incorporate an additional supervised loss $\mathcal{L}_{sup}$. This loss directs the learning process for the entire Robult network and adapts based on the task type: $L_1$ loss for regression tasks and cross-entropy $\mathcal{L}_{ce}$ for classification tasks.
A detailed training procedure, including loss formulations and implementation details, is provided in Appendix \ref{sec:app_training}.
\section{Experimental Results}
\label{sec:experiment}
\subsection{Datasets and Metrics}
    \textbf{Dataset:} We conduct experiments on the following datasets.
\begin{itemize}
    \item
    \textit{CMU-MOSI} \cite{mosi} \& \textit{CMU-MOSEI} \cite{mosei}: Containing three modalities - text, audio, and visual, supporting sentiment analysis and emotion recognition tasks. Each video is labeled on a scale from -3 (negative) to 3 (positive) sentiment.
    \item 
    \textit{MM-IMDb} \cite{mmimdb}: Containing image and text modalities, serving genre classification task - which involves multi-label classification as a movie has several genres.
    \item 
    \textit{UPMC Food-101} \cite{food101}: Containing two modalities, text and images, this dataset is a classification dataset consisting of $101$ food categories.
    \item 
    \textit{Hateful Memes} \cite{hatememes}: Containing text and image modalities, this dataset aims at identifying hate speech in memes. This dataset includes challenging examples that are similar to hateful ones but are actually harmless.
\end{itemize}

\textbf{Metrics:} For sentiment analysis related to CMU-MOSI and CMU-MOSEI datasets, we adopt mean absolute error (MAE), correlation (Cor), binary accuracy, and F1 score, following \cite{icml22,mosi_2}. Here, binary categories determine positive sentiment scores $(> 0)$ or negative ones $(< 0)$. 
For the evaluation of the three remaining datasets, we adhere to the metrics specified in \cite{cvpr23_2}. With the MM-IMDb dataset, the multi-label classification performance is assessed using F1-Macro. The classification accuracy is employed for the UPMC Food-101 dataset. Lastly, for Hateful Memes, the evaluation is based on the AUROC metric.

\subsection{Baselines and Experimental Settings}
\textbf{Baselines:} 
We incorporate several state-of-the-art approaches representing popular strategies into our comparative evaluation. Specifically, GMC \cite{icml22} serves as a contrastive learning-based approach, ActionMAE \cite{aaai23} represents a generation-based method, and we include a Transformer-based approach proposed in \cite{cvpr23_2}, referred to as Prompt-Trans for brevity. To ensure optimal reproducibility, we inherit the implementations of all baseline methods from their original code bases. Additionally, we implement unimodal frameworks (Unimodal) for each modality, trained in a supervised manner with available labels, to serve as our baseline comparison.

\textbf{Implementation details:} 
To ensure a fair comparison, we use similar encoder architectures for processing raw data modalities whenever possible. The unimodal baselines are designed with the same architectures as Robult, each with its own classifier. 
For Robult, positive samples for the soft P-U loss are determined after discretizing labels if needed. Specifically, in the cases of CMU-MOSI and CMU-MOSEI datasets, label information in the range of $[-3, 3]$ is quantified into $7$ discrete categories ($-3, -2, \dots, 3$). Additionally, for the multi-label dataset MM-IMDb, two samples are considered positive if they share all the same labels. 
Regarding Prompt-Trans, we only report its results for three datasets involving two modalities (MM-IMDb dataset, UPMC Food-101 dataset, and Hateful Memes dataset), as the extension to multiple modalities cannot be directly inferred from the original work \cite{cvpr23_2}.

\textbf{Experimental details:} 
The primary focus of our performance reporting is on two extreme scenarios: semi-supervised settings with only $5\%$ labeled data and scenarios where only a single modality is presented during evaluation. All reported results are averaged over 3 different random seeds. In the semi-supervised setup, the newly created labeled portion is ensured to maintain the correct label ratio as the original training sets. Additional experiments extending these two settings to more modalities and a higher percent of labeled data are detailed in Appendix \ref{sec:app_missing} and \ref{sec:app_semi} respectively.
% For the Sentiment Analysis task using the CMU-MOSI and CMU-MOSEI datasets, we adhere to the settings outlined in \cite{icml22}, which involve temporally-aligned versions of these datasets generated by \cite{mosi_mosei_align}.
% For the classification tasks associated with the MM-IMDb, UPMC Food-101, and Hateful Memes datasets, we initially generate text and visual embeddings offline using a pretrained ViLT framework \cite{vilt}. Subsequently, all models are trained using these embeddings. This specific procedure is intentionally conducted to ensure fair evaluation, as Prompt-Trans \cite{cvpr23_2} also involves the same frozen ViLT framework in their training process. We follow Prompt-Trans \cite{cvpr23_2} for the preprocessing procedures of raw texts and images.
Specific details on implementation settings relating to each dataset are provided in Appendix - \ref{sec:app_implement}.

\subsection{Main Quantitative Results}
\begin{table}[!t]
% \begin{minipage}{0.52\linewidth}
\centering
\setlength\tabcolsep{3pt}
\resizebox{\columnwidth}{!}{%
\begin{tabular}{@{}lcccc|cccc@{}}
% \toprule
                                                     & \multicolumn{4}{c}{CMU-MOSI}                                                                                          & \multicolumn{4}{c}{CMU-MOSEI}                                                                      \\ \midrule
 Metrics & Unimodal                     & GMC                          & ActionMAE                    & \textbf{Robult}                       & Unimodal                     & GMC                          & ActionMAE & \textbf{Robult}                       \\ \midrule
\multicolumn{8}{l}{\textit{Text Modality:}} \\
 MAE ($\downarrow$)                       & 1.41                         & \textcolor{blue}{1.407} & 1.476                        & \textcolor{red}{\textbf{1.397}} & \textcolor{blue}{0.81}  & 0.815                        & 1.115     & \textcolor{red}{\textbf{0.784}} \\
                           Corr ($\uparrow$)                      & 0.137                        & \textcolor{blue}{0.14}  & 0.066                        & \textcolor{red}{\textbf{0.144}} & \textcolor{blue}{0.383} & 0.346                        & 0.136     & \textcolor{red}{\textbf{0.459}} \\
                           F1 ($\uparrow$)                        & 0.551                        & \textcolor{blue}{0.559} & 0.535                        & \textcolor{red}{\textbf{0.578}} & \textcolor{blue}{0.717} & 0.716                        & 0.614     & \textcolor{red}{\textbf{0.739}} \\
                           Acc ($\uparrow$)                       & 0.553                        & \textcolor{blue}{0.562} & 0.47                         & \textcolor{red}{\textbf{0.569}} & \textcolor{blue}{0.712} & 0.708                        & 0.603     & \textcolor{red}{\textbf{0.732}} \\ \midrule
\multicolumn{8}{l}{\textit{Audio Modality:}} \\                      
MAE ($\downarrow$)                       & 1.576                        & \textcolor{blue}{1.518} & 1.546                        & \textcolor{red}{\textbf{1.415}} & 0.842                        & \textcolor{blue}{0.836} & 1.215     & \textcolor{red}{\textbf{0.825}} \\
                           Corr ($\uparrow$)                      & 0.041                        & -0.065                       & \textcolor{blue}{0.046} & \textcolor{red}{\textbf{0.085}} & 0.111                        & \textcolor{blue}{0.193} & 0.101     & \textcolor{red}{\textbf{0.221}} \\
                           F1 ($\uparrow$)                        & \textcolor{blue}{0.512} & 0.457                        & 0.508                        & \textcolor{red}{\textbf{0.539}} & 0.618                        & \textcolor{blue}{0.642} & 0.634     & \textcolor{red}{\textbf{0.679}} \\
                           Acc ($\uparrow$)                       & \textcolor{blue}{0.496} & 0.46                         & 0.467                        & \textcolor{red}{\textbf{0.535}} & 0.599                        & \textcolor{blue}{0.63}  & 0.543     & \textcolor{red}{\textbf{0.65}}  \\ \midrule
\multicolumn{8}{l}{\textit{Vision Modality:}} \\
MAE ($\downarrow$)                       & \textcolor{blue}{1.451} & 1.497                        & 1.511                        & \textcolor{red}{\textbf{1.425}} & 0.891                        & \textcolor{blue}{0.839} & 1.127     & \textcolor{red}{\textbf{0.826}} \\
                           Corr ($\uparrow$)                      & \textcolor{blue}{0.044} & -0.07                        & -0.03                        & \textcolor{red}{\textbf{0.086}} & 0.163                        & \textcolor{blue}{0.2}   & 0.104     & \textcolor{red}{\textbf{0.201}} \\
                           F1 ($\uparrow$)                        & \textcolor{blue}{0.585} & 0.446                        & 0.511                        & \textcolor{red}{\textbf{0.593}} & \textcolor{blue}{0.637} & 0.621                        & 0.594     & \textcolor{red}{\textbf{0.647}} \\
                           Acc ($\uparrow$)                       & 0.425                        & 0.449                        & \textcolor{blue}{0.514} & \textcolor{red}{\textbf{0.522}} & \textcolor{blue}{0.624} & 0.62                         & 0.561     & \textcolor{red}{\textbf{0.632}} \\ \midrule
\multicolumn{8}{l}{\textit{Full Modality:}} \\
MAE ($\downarrow$)                       & \textcolor{blue}{1.394} & 1.47                         & 1.496                        & \textcolor{red}{\textbf{1.392}} & \textcolor{blue}{0.783} & 0.819                        & 1.103     & \textcolor{red}{\textbf{0.779}} \\
                           Corr ($\uparrow$)                      & \textcolor{blue}{0.186} & 0.101                        & -0.092                       & \textcolor{red}{\textbf{0.247}} & \textcolor{blue}{0.364} & 0.328                        & 0.337     & \textcolor{red}{\textbf{0.504}} \\
                           F1 ($\uparrow$)                        & \textcolor{blue}{0.597} & 0.497                        & 0.553                        & \textcolor{red}{\textbf{0.657}} & \textcolor{blue}{0.73}  & 0.693                        & 0.694     & \textcolor{red}{\textbf{0.744}} \\
                           Acc ($\uparrow$)                       & \textcolor{blue}{0.594} & 0.498                        & 0.477                        & \textcolor{red}{\textbf{0.63}}  & \textcolor{blue}{0.729} & 0.688                        & 0.643     & \textcolor{red}{\textbf{0.741}} \\ \bottomrule
\end{tabular}
}
\caption{Results on CMU-MOSI, CMU-MOSEI.}
\label{tab:mosi_mosei}
\end{table}
All results are shown in tables with the best outcomes in \textcolor{red}{\textbf{red}} and the second-best in \textcolor{blue}{blue}. 

\textbf{Sentiment Analysis:} The results for CMU-MOSI and CMU-MOSEI datasets are summarized in Table \ref{tab:mosi_mosei}. For both datasets, Robult significantly outperforms all the compared methods, suggesting its effectiveness and consistency in semi-supervised and missing modality scenarios. Regarding CMU-MOSI, due to its smaller scale compared to CMU-MOSEI, the labeled portions are also smaller. This condition poses a challenge for existing baselines that heavily rely on label information. In contrast, Robult effectively addresses this challenge, demonstrating the ability to extract meaningful representations even with limited labeled data. On CMU-MOSEI, Robult consistently produces superior representations, achieving the best performances across all recorded metrics. Notably, Robult improves the correlation (Corr) between the predicted sentiment levels and ground truth by up to $19.8\%$, outperforming the second-best method, which is the unimodal for textual data.

\begin{table}[]
    \begin{minipage}{\columnwidth}
\setlength\tabcolsep{3pt}
\resizebox{\textwidth}{!}{%
\begin{tabular}{@{}lccccc@{}}
\toprule
  & Unimodal & Prompt-Trans                          & GMC                          & ActionMAE & \textbf{Robult}                                \\ \midrule
 & \multicolumn{5}{l}{\textit{MM-IMDb - F1 Macro ($\uparrow$):}}                                                                                             \\
Text      & 0.24     & 0.198                                 & \textcolor{blue}{0.296} & 0.055     & \textcolor{red}{\textbf{0.321}} \\
Image     & 0.207    & 0.148                                 & \textcolor{blue}{0.291} & 0.039     & \textcolor{red}{\textbf{0.298}} \\
Full      & 0.196    & 0.268                                 & \textcolor{blue}{0.307} & 0.171     & \textcolor{red}{\textbf{0.332}} \\ \midrule
& \multicolumn{5}{l}{\textit{UPMC Food-101 - Accuracy ($\uparrow$):}}                                                                                       \\
Text      & 0.321    & 0.151                                 & \textcolor{blue}{0.395} & 0.196     & \textcolor{red}{\textbf{0.435}} \\
Image     & 0.296    & 0.111                                 & \textcolor{blue}{0.382} & 0.132     & \textcolor{red}{\textbf{0.415}} \\
Full      & 0.138    & \textcolor{blue}{0.432}          & 0.41                         & 0.358     & \textcolor{red}{\textbf{0.446}} \\ \midrule
 & \multicolumn{5}{l}{\textit{Hateful Memes - AUROC ($\uparrow$):}}                                                                                          \\
Text      & 0.584    & 0.511                                 & \textcolor{blue}{0.617} & 0.528     & \textcolor{red}{\textbf{0.623}} \\
Image     & 0.524    & 0.475                                 & \textcolor{blue}{0.528} & 0.508     & \textcolor{red}{\textbf{0.596}} \\
Full      & 0.618    & \textcolor{red}{\textbf{0.635}} & 0.616                        & 0.542     & \textcolor{blue}{0.632}          \\ \bottomrule
\end{tabular}
}
\caption{Results on MM-IMDb, UPMC Food-101, Hateful Memes.}
\label{tab:cls}
\end{minipage}
\begin{minipage}{\columnwidth}
        \centering
        \includegraphics[width=\textwidth]{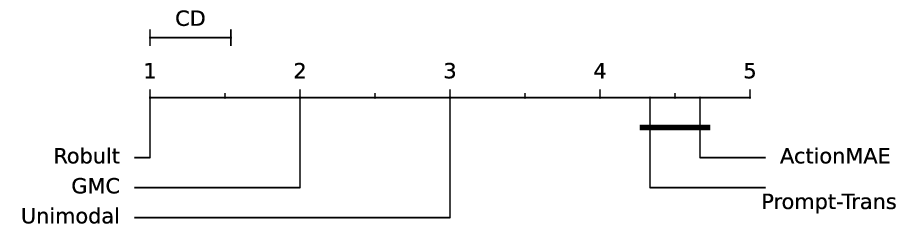}
        \captionof{figure}{CD diagram showing the mean rank of each method on three datasets.}
        \label{fig:cd_diag}
\end{minipage}
\end{table}
\textbf{Classification tasks:} In Table \ref{tab:cls}, empirical results for three classification tasks show that Robult consistently outperforms existing approaches and baselines in most cases, except for one scenario on the Hateful Memes dataset with the full modality available, where Robult achieves comparable performance with Prompt-Trans \cite{cvpr23_2}. Notably, the Hateful Memes dataset includes samples with ``benign confounders'', negatively impacting performance when models rely solely on single modalities \cite{hatememes}. Leveraging the soft Positive-Unlabelled loss, Robult effectively addresses and mitigates performance gaps with either single modality inputs or the full ones.
In addition, we calculate F1 macro scores for all methods on these three datasets in the unimodal and multimodal cases. We further visualize a Critical Difference Diagram \cite{cd_diagram} in Figure \ref{fig:cd_diag}. This diagram visually represents the performance among different machine learning algorithms across various datasets by displaying the mean performance ranks, with lower being better, and connecting statistically indistinguishable groups (within $95\%$ confidence level) with a thin horizontal bar, as per the Friedman hypothesis test. From the diagram, Robult exhibits a clear improvement gap compared to other state-of-the-art methods in average ranks, while ActionMAE and Prompt-Trans show no statistically significant difference in their performance.
% Quantitatively, across two main tasks and five datasets, the Robult pipeline demonstrates its superiority over existing approaches. 
% In the following section, we will further assess the qualitative results of Robult and conduct ablation studies to evaluate the impact of each module to the overall performance.

\subsection{Main ablation studies}
% In this section, we provide two key analyses: an evaluation of the quality of the learned representations and the main ablation study of Robult. 
Owing to space constraint, we provide here a key analysis of ablation study for different components of Robult. 
For a more comprehensive analysis, please refer to the additional experiments in Appendix \ref{sec:app_exp}, which offer further insights into how architectural choices, Soft-PU loss, and weighting schemes influence Robult's performance quantitatively and qualitatively.

\begin{table}[!th]
\centering
\resizebox{\columnwidth}{!}{%
\begin{tabular}{@{}rcccccc@{}}
\toprule
Metrics  & \multicolumn{1}{c}{GMC}      & \multicolumn{1}{c}{\textbf{Robult}}   & \multicolumn{1}{l}{Robult (1)} & \multicolumn{1}{l}{Robult (2)} & \multicolumn{1}{l}{Robult (3)}  & \multicolumn{1}{l}{Robult (4)} \\ \midrule
         & \multicolumn{6}{l}{\textit{CMU-MOSI - Text Modality:}}                                                                                                                                                                               \\
MAE      & \textcolor{blue}{1.407} & \textcolor{red}{\textbf{1.397}} & 1.589                                 & 1.511                                 & 1.443                                 & 1.429                                 \\
Corr     & \textcolor{blue}{0.14}  & \textcolor{red}{\textbf{0.144}}                        & 0.047                                 & 0.101                                 & 0.051                                 & 0.123                                 \\
F1       & 0.559                        & \textcolor{blue}{0.578}                                 & 0.542                                 & 0.52                                  & \textcolor{red}{\textbf{0.593}} & 0.571                                 \\
Acc      & 0.562                        & \textcolor{blue}{0.569}          & 0.544                                 & 0.523                                 & 0.422                                 & \textcolor{red}{\textbf{0.573}} \\ \midrule
         & \multicolumn{6}{l}{\textit{CMU-MOSI - Audio Modality:}}                                                                                                                                                                              \\
MAE      & 1.518                        & \textcolor{red}{\textbf{1.415}} & 1.586                                 & 1.561                                 & \textcolor{blue}{1.494}          & 1.495                                 \\
Corr     & -0.065                       & \textcolor{red}{\textbf{0.085}} & 0.023                                 & 0.005                                 & 0.046                                 & \textbf{\textcolor{red}{0.085}}          \\
F1       & 0.457                        & \textcolor{red}{\textbf{0.539}} & \textcolor{blue}{0.526}          & 0.499                                 & 0.51                                  & 0.517                                 \\
Acc      & 0.46                         & \textcolor{red}{\textbf{0.535}} & \textcolor{blue}{0.518}          & 0.502                                 & 0.442                                 & 0.509                                 \\ \midrule
         & \multicolumn{6}{l}{\textit{CMU-MOSI - Vision Modality:}}                                                                                                                                                                                      \\
MAE      & 1.497                        & \textcolor{red}{\textbf{1.425}} & 1.663                                 & 1.711                                 & \textcolor{blue}{1.445}          & 1.504                                 \\
Corr     & -0.07                        & \textcolor{red}{\textbf{0.086}} & 0.025                                 & -0.023                                & \textcolor{blue}{0.041}          & -0.066                                \\
F1       & 0.446                        & \textcolor{red}{\textbf{0.593}} & 0.519                                 & 0.485                                 & \textcolor{blue}{0.571}          & 0.459                                 \\
Acc      & 0.449                        & \textcolor{red}{\textbf{0.522}} & \textcolor{blue}{0.519}          & 0.465                                 & 0.47                                  & 0.448                                 \\ \midrule
         & \multicolumn{6}{l}{\textit{CMU-MOSI - Full Modality:}}                                                                                                                                                                                        \\
MAE      & 1.47                         & \textcolor{red}{\textbf{1.392}}                        & 1.588                                 & 1.434                                 & \textcolor{blue}{1.411}          & 1.459                                 \\
Corr     & 0.101                        & \textcolor{red}{\textbf{0.247}} & 0.071                                 & \textcolor{blue}{0.239}          & 0.166                                 & 0.229                                 \\
F1       & 0.497                        & \textcolor{red}{\textbf{0.657}} & 0.524                                 & 0.567                                 & 0.549                                 & \textcolor{blue}{0.6}            \\
Acc      & 0.498                        & \textcolor{red}{\textbf{0.63}}  & 0.523                                 & 0.566                                 & 0.552                                 & \textcolor{blue}{0.601}          \\ \midrule
         & \multicolumn{6}{l}{\textit{Hateful Memes - Text Modality:}}                                                                                                                                                                                   \\
AUROC    & \textcolor{blue}{0.617} & \textcolor{red}{\textbf{0.623}} & 0.528                                 & 0.59                                  & 0.605                                 & 0.576                                 \\
Acc & \textcolor{blue}{0.581} & \textcolor{red}{\textbf{0.59}}  & 0.535                                 & 0.556                                 & 0.571                                 & 0.562                                 \\ \midrule
         & \multicolumn{6}{l}{\textit{Hateful Memes - Image Modality:}}                                                                                                                                                                                  \\
AUROC    & 0.528                        & \textcolor{red}{\textbf{0.596}} & 0.518                                 & 0.582                                 & \textcolor{blue}{0.588}          & 0.566                                 \\
Acc & \textcolor{blue}{0.551} & \textcolor{red}{\textbf{0.562}} & 0.524                                 & \textcolor{blue}{0.551}          & 0.526                                 & 0.539                                 \\ \midrule
         & \multicolumn{6}{l}{\textit{Hateful Memes - Full Modality:}}                                                                                                                                                                                   \\
AUROC    & 0.616                        & \textcolor{blue}{0.632}          & 0.538                                 & 0.618                                 & \textcolor{red}{\textbf{0.634}} & 0.582                                 \\
Acc & 0.532                        & \textcolor{red}{\textbf{0.595}} & 0.542                                 & 0.55                                  & \textcolor{blue}{0.554}          & 0.552                                 \\ \bottomrule
\end{tabular}
}
\caption{Ablation analysis on CMU-MOSI and Hateful Memes datasets for Robult.}
\label{tab:ablation}
\end{table}
We evaluate the impact of each loss component on Robult's performance using Hateful Memes dataset, which mirror the semi-supervised and missing modalities conditions of our main experiments. This analysis involves testing variations of Robult with different ablations.
(1) \textit{Removal of $\mathcal{L}_{sup}$} - this setting utilizes available label information only in $\mathcal{L}_{(u)lb}$, so Robult can only produce latent representations. An additional Logistic Regressor is trained with these representations as its input, and this pipeline's final scores are reported. (2) \textit{Removal of $\mathcal{L}_{rec}$} - this setting discards $\mathcal{L}_{rec}$, corresponding to our Objective \ref{obj:entropy}. (3) \textit{Removal of $\mathcal{L}_{lb}$} - this setting makes the learning of Objective \ref{obj:mi} rely only on $\mathcal{L}_{ulb}$. (4) \textit{Removal of $\mathcal{L}_{ulb}$} - this setting associates Objective \ref{obj:mi} exclusively with $\mathcal{L}_{lb}$.
Table \ref{tab:ablation} summarizes the results of this ablation experiment. Overall, any ablation negatively impacts the performance of Robult. In particular, the absence of $\mathcal{L}_{sup}$ significantly worsens the performance, as there is no loss guiding the learning of Robult's classifier, which is crucial for generating soft label information consumed by the soft Positive-Unlabeled loss $\mathcal{L}_{ulb}$. Consequently, this ablation adversely affects two loss components, explaining the poorest result among all variations. The removal of $\mathcal{L}_{rec}$ particularly harms the performance with unimodal inputs, aligning with the motivation for Objective \ref{obj:entropy}, as the unique information $U^i$ is no longer preserved. In two remaining cases, both ablations diminish Robult's overall performance, indicating their equal contribution to achieving Objective \ref{obj:mi}. 
\section{Contributions \& Limitations}
\label{sec:conclusion}

\textbf{Contributions:} 
Our Robult pipeline leverages limited label data through a soft Positive-Unlabelled (PU) loss and latent reconstruction loss, enhancing modality interactions and preserving unimodal data integrity. It supports various modality types and quantities, scales linearly with modalities, and functions independently of specific architectures. This flexibility facilitates integration with existing DL frameworks, advancing multimodal learning in practical settings.

\textbf{Limitations.}
Robult's design presumes that the proximity of positive couplets follows a Gaussian distribution, a method proven empirically but not theoretically. Future work should seek theoretical validation for this assumption. Moreover, with our setting, the potential of labeled data in scenarios with missing modalities in training remains untapped. Exploring these cases could further improve Robult's effectiveness in complex real-world applications.

%
% ---- Bibliography ----

% \input{sections/reproducability}

\newpage
\section*{Acknowledgements}
The work of Duy A. Nguyen was supported in part by a PhD fellowship from the VinUni-Illinois Smart Health Center, VinUniversity, Hanoi, Vietnam.
\bibliographystyle{named}
\bibliography{references}

\newpage

\setcounter{page}{1}
\appendix
\section{Related Works}
\label{sec:literature}

% While the exploration of diverse data modalities to enhance task performance has been extensively studied \cite{cvpr21_2, icml23, cvpr23}, the application of multimodal frameworks in real configurations is constrained by their non-ideal settings. Several studies aim to address these challenges.

\textbf{Semi-supervised Multimodal Learning:} 
Several works acknowledge the challenge of fully labeled datasets in the multimodal literature and provide targeted solutions for specific applications \cite{semi_1,semi_4,semi_5}. For instance, in \cite{semi_1}, the authors tackle the semi-supervised scenario in scene text recognition by enforcing consistency between weakly augmented pseudo-labels and strongly augmented views. 
% SMIN \cite{semi_2} addresses conversational emotion recognition tasks through intra-modal and cross-modal interactive modules inspired by auto-encoders. 
% In \cite{semi_3}, labeled hash codes are learned using label signals, preserving the data structure of unlabeled ones, followed by importance differentiation regression for final multimodal hashing. 
Authors in \cite{semi_4} propose an area-similarity contrastive loss for medical image segmentation, leveraging cross-modal information to enhance representations of unlabeled data. 
Liang et al. \cite{semi_5} derive two lower bounds of multimodal interaction from an information-theoretic perspective, applicable for pre-analysis of multimodal interaction effects. 
A common technique in general semi-supervised learning is loss reweighting based on pseudolabel uncertainty, similar to our PU loss. 
These methods aim to mitigate confirmation bias and are widely used in unsupervised domain adaptation \cite{loss_weighting_1}, particularly in image processing applications \cite{loss_weighting_image_2}. 
However, these efforts primarily focus on semi-supervised scenarios in specific tasks and certain modalities (e.g., text-images), limiting their applicability to general cases.

\textbf{Missing modalities:}
Many multimodal fusion methods rely on a complete set of modalities, but deployment settings often lack such ideal conditions, leading to adverse effects when using these strategies \cite{kdd20,cvpr22}. To address this challenge, some approaches aim to create models resilient to missing modalities \cite{aaai_21,cvpr22,icml22,aaai23,cvpr23_2}. For instance, Wang et al. \cite{kdd20} optimize training by considering incomplete data samples to generate unimodal teachers guiding a multimodal student. Smil \cite{aaai_21} approximates latent features of modality-incomplete data using Bayesian meta-learning. GMC \cite{icml22} preserves geometric alignment in multimodal representations, enabling unimodal representations to substitute for absent representations of other modalities. ActionMAE, inspired by the masked autoencoder idea \cite{nips22}, learns to predict the latent representation of a missing modality by randomly dropping its feature token and learning to reconstruct it. Despite success in certain scenarios, these frameworks often rely on labeled signals, implicitly or explicitly, in the training dataset, limiting their general applicability.

% Our study diverges from these works by proposing a versatile framework proficient in both semi-supervised scenarios and scenarios with missing modalities. Accounting for this gap in the literature, we aim to provide a universally applicable solution by addressing both challenges simultaneously, while maintaining information-rich representations. 
\section{Robult supplementary details}
\subsection{Minimizing Conditional Entropy with Latent
Reconstruction Error}
\label{sec:sup_entropy} 
\begin{figure}[ht]
    \centering
    \includegraphics[width=\columnwidth]{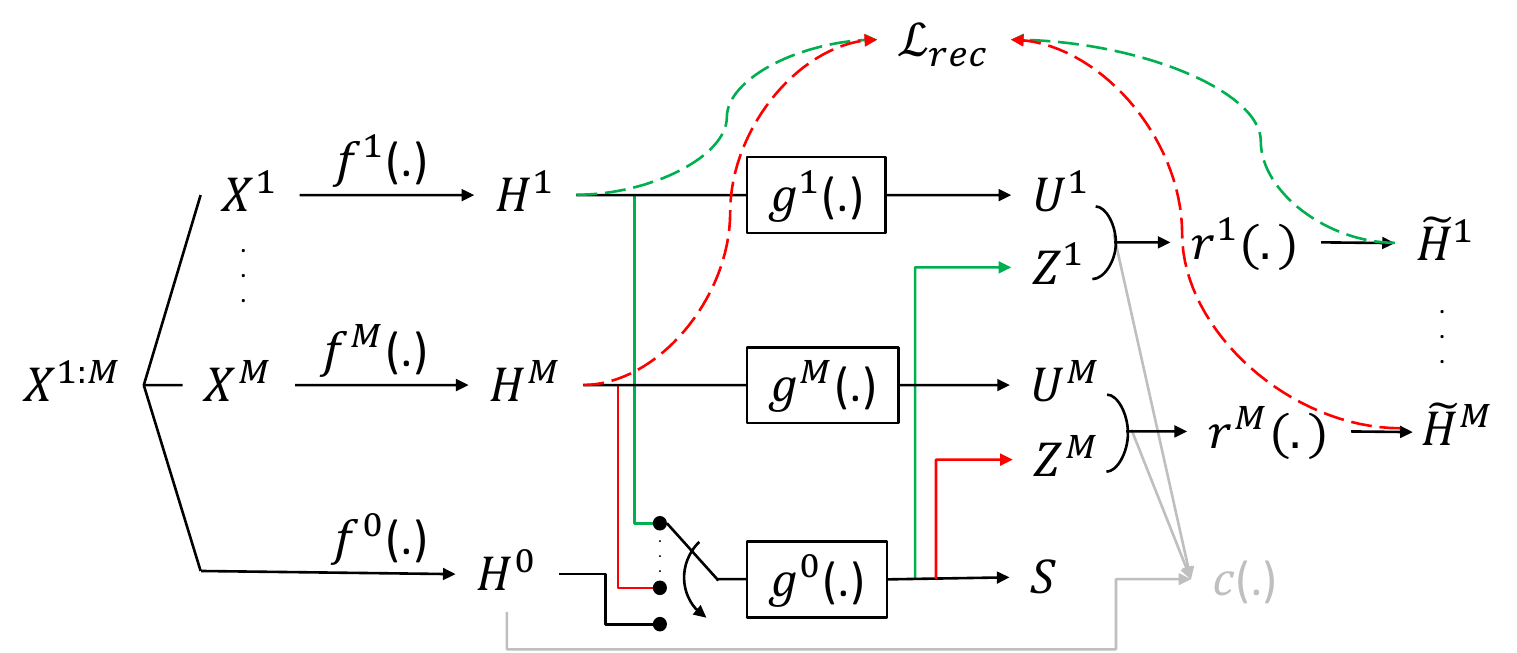}
    \caption{Reconstruction procedure of Robult. This procedure only applied in training stage.}
    \label{fig:reconstruct}
\end{figure}

As explained in the primary text, our approach to achieve Objective \ref{obj:entropy} involves a reconstruction procedure with two components: the reconstruction module $r^i(U^i, Z^i) = \Tilde{H}^i$ and the latent reconstruction loss $\mathcal{L}_{rec}$. This procedure is illustrated in Figure \ref{fig:reconstruct}.
It is important to note that these reconstruction modules $r^i(.)$ are used exclusively during the learning process to optimize individual branches $g^i(.)$, incurring no additional overhead during the evaluation or deployment stages. As this reconstruction is carried out in the latent space, the module $r^i(.)$ can be uniformly designed, irrespective of the characteristics of input modalities. In our Robult design, $r^i(.)$ is simply a two-layered MLP with ReLU activation in the middle, applied across all five datasets.

\subsection{Maximizing Mutual Information with Soft Positive-Unlabeled Contrastive Learning}
\label{sec:sup_mi}
In this section, we would derive the lower bound of mutual information between fused latent $S$ and unimodal representation $Z^i$ as a Positive-Unlabelled learning objective, which relax the assumption about full presence of labels in training dataset. This derivation explains Result \ref{eq:res_1} in the main manuscript.

The ultimate goal is to maximize the following MI quantity: 
\begin{equation*}
    \mathcal{I}(S, Z^i) = d_{KL}\left(p_{S, Z^i} || p_S \otimes p_{Z^i}\right)
\end{equation*}
This essentially means that the KL divergence between the joint distribution $p_{S, Z^i}$ and the product of marginal distribution $p_S \otimes p_{Z^i}$ should be maximized. As defined in the main manuscript, $F$ is the flag indicator denotes whether a couplet $(s, z^i)$ is sampled from the joint distribution $p_{S,Z^i}$ $(F=1)$ or from product of marginal distribution $p_S \otimes p_{Z^i}$ $(F=0)$:
% In addition, $L$ is a variable denoting whether the pair $(s, z^i)$ is labelled $(L=1)$ or not $(L=0)$. Within a mini-batch of size $b$, we have:
\begin{equation}
\begin{aligned}
    &p(S,Z^i | F=1) = p_{S,Z^i}; \hspace{0.5em} p(S,Z^i | F=0) = p_S \otimes p_{Z^i}; 
    % \\
    % &p(L=1) = \frac{b_l}{\binom{b}{2}}; \hspace{0.5em} p(L=0) = \frac{\binom{b}{2} - b_l}{\binom{b}{2}}; \\
    % &p(F=1 | L=1) = \frac{b_{lp}}{b_l}; \hspace{0.5em} p(F=0 | L=1) = \frac{b_{ln}}{b_l}. \\
\end{aligned}
\label{eq:priors}
\end{equation}
% In Eq. \ref{eq:priors}, binomial coefficient $\Tilde{b} = \binom{b}{2}$; $b_l$ is the number of labelled couplets within mini-batch $b$; $b_{lp}$ and $b_{ln}$ are labelled positive and negative pairs, respectively. 

Applying Bayes' rule, the posterior for $F = 1$ is given by:
\begin{equation}
\begin{split}
    &p(F = 1| S, Z^i) = \frac{p(S, Z^i | F=1)p(F=1)}{p(S, Z^i)} \\
    &= \frac{p(S, Z^i | F=1)p(F=1)}{p(S, Z^i | F=1)p(F=1) + p(S, Z^i | F=0)p(F=0)} \\
    &= \frac{p_{S,Z^i} \cdot p(F=1)}{p_{S,Z^i} \cdot p(F=1) + p_S \otimes p_{Z^i} \cdot p(F=0)}.
\end{split}
\label{eq:posterior_c}
\end{equation}
Applying logarithm operation on both side of Equation \ref{eq:posterior_c}:
\begin{equation}
\begin{split}
    \mbox{log}p(F = 1| S, Z^i) &= -\mbox{log}\left( 1 + k \frac{p_S \otimes p_{Z^i}}{p_{S,Z^i}} \right) \\
    &\leq -\mbox{log}k + \mbox{log}\frac{p_{S,Z^i}}{p_S \otimes p_{Z^i}}, \\
\end{split}
\end{equation}
in which
\begin{equation}
    k = \frac{p(F=0)}{p(F=1)}.
\end{equation}

Taking expectation w.r.t $p_{S,Z^i}$ (or $p(S, Z^i | F=1)$), we can bound the mutual information as
\begin{equation}
    \mathcal{I}(S,Z^i) \geq \mathbb{E}_{p(S, Z^i | F=1)} \mbox{log}p(F=1| S, Z^i) + \mbox{log}k
\end{equation}
Here, the true distribution $p(F = 1| S, Z^i)$ is unknown, so we approximate it with a well-established non-parametric model $v: S \times Z^i \rightarrow [0,1]$ \cite{cvpr21,cvpr23}:
\begin{equation}
\begin{split}
    &\mathcal{I}(S,Z^i) \ge \mathbb{E}_{p(S, Z^i | F=1)}\mbox{log}v(S,Z^i) + \mbox{log}k \\
    \text{where} \\
    &v(s_j,z^i_k) = \frac{\phi(s_j, z_k^i)}{\sum_h^B\phi(s_j, z_h^i)}; \\
    &\phi(s_j, z_k^i) = exp(<s_j; z_k^i> / \tau).
\end{split}
\label{eq:h_func}
\end{equation}
In addition, let $c$ be the number of underlying classes and assume the labels are uniformly distributed, we have the probability that a couplet is sharing a label is $p(f_k=1) = \frac{1}{c^2}$. Within mini-batch of size $B$, consider the scenario in which the number of positive couplets $B_p$ is greater than the number of negative ones $B_n$ (hence $B_p > \frac{\binom{B}{2}}{2} = \frac{\Tilde{B}}{2}$): 
\begin{equation*}
\begin{aligned}
    p(B_p) &= \binom{\Tilde{B}}{B_p} \cdot \frac{1}{c^{2B_p}} \\
            &\le \binom{\Tilde{B}}{\frac{\Tilde{B}}{2}} \cdot \frac{1}{c^{\Tilde{B}}}
\end{aligned}
\end{equation*}
It should be noted that this possibility $p(B_p)$ is upper-bounded by a small quantity given $B>1$ and $c\ge2$ (smaller than $0.1$ in case $B=8$ and $c=2$), and get smaller when $B$ and $c$ increase. Intuitively, the possibility that the positive couplets outnumber the negative ones is negligible, hence, it normally hold true that:
$$
\log k = \log \frac{p(F=0)}{p(F=1)} \ge 0.
$$
With this realization, Result \ref{eq:res_1} can be derived from Eq. \ref{eq:h_func} as follow:
\begin{equation}
    \begin{aligned}
        \mathcal{I}(S,Z^i) &\ge \mathbb{E}_{p(S, Z^i | F=1)}\mbox{log}v(S,Z^i) + \mbox{log}k \\
        &\ge \mathbb{E}_{p(S, Z^i | F=1)}\mbox{log}v(S,Z^i) \\
        &= \mathbb{E}_{p_{S, Z^i}}\mbox{log}v(S,Z^i).
        % &= \mathbb{E}_{p(S, Z^i | F=1, L=1)}\mbox{log}h(S,Z^i) + \mathbb{E}_{p(S, Z^i | F=1, L=0)}\mbox{log}h(S,Z^i) \\
    \end{aligned}
\end{equation}
\subsection{Training Strategy}
\label{sec:app_training}
We employ an end-to-end training pipeline that can process two objectives \ref{obj:entropy} and \ref{obj:mi} independently, as demonstrated by Algorithm \ref{alg:training}. 
In general, we selectively perform gradient calculations on different modules of Robult based on the specific losses. 
This selection process offers dual benefits: (1) minimal processing overhead with a single forward pass, and (2) effective restriction of the losses' impact only on their target modules.
\begin{algorithm}[!ht]
   \caption{Robult training strategy}
   \label{alg:training}
\begin{algorithmic}
   \STATE {\bfseries Input:} \\
   \STATE {$\triangleright \hspace{0.2cm}$ Training dataset $\mathcal{D}_{train}$}
   \STATE {$\triangleright \hspace{0.2cm}$ Robult framework $\mathcal{RB}$}
   \STATE {$\triangleright \hspace{0.2cm}$ Optimizer $\mathcal{O}$}
   
   \FUNCTION {\textit{ParametersToggle}($f$: \textit{flag\_variable})}
   \IF{$f = 0$}
   \STATE {Toggle all $\mathcal{RB}$ parameters to require gradient calculation}
   \ELSIF{$f = 1$}
   \STATE {Toggle all $\mathcal{RB}$ parameters to \textbf{NOT} require gradient calculation}
   \STATE {Toggle all $g^i(.)$ parameters $(i=1,\dots,M)$ to require gradient calculation }
   \ELSIF{$f = 2$}
   \STATE {Toggle all $\mathcal{RB}$ parameters to \textbf{NOT} require gradient calculation}
   \STATE {Toggle all $f^i(.), g^i(.)$ parameters $(i=\overline{0,M})$ to require gradient calculation}
   \ENDIF
   \ENDFUNCTION
   
   \FOR{$B_i; Y_i$ {\bfseries in} $\mathcal{D}_{train}$}
   \STATE {$\triangleright$ \textit{single forward pass}}
   \STATE $\Tilde{Y}_i, H_i, Z_i, U_i, S = \mathcal{RB}(B_i)$ 
   \STATE {$\triangleright$ \textit{loss calculations}}
   \STATE $l_{cls} = \mathcal{L}_{cls}(\Tilde{Y}_i, Y_i)$ 
   \STATE $l_{rec} = \mathcal{L}_{rec}(H_i, Z_i, U_i)$ 
   \STATE $l_{(u)lb} = \mathcal{L}_{(u)lb}(Z_i, S)$ 
   \STATE {$\triangleright$ \textit{gradient calculations}}
   \STATE {Call \textit{ParametersToggle}($f=1$); backward with $l_{rec}$}
   \STATE {Call \textit{ParametersToggle}($f=2$); backward with $l_{lb}$ and $l_{ulb}$}
   \STATE {Call \textit{ParametersToggle}($f=0$); backward with $l_{cls}$}
   \STATE {$\triangleright$ \textit{single backward pass}}
   \STATE {Optimizer $\mathcal{O}$ update $\mathcal{RB}$ parameters with above gradient infomation}
   \ENDFOR
\end{algorithmic}
\end{algorithm}

\subsection{Robult's Complexity Analysis}
Robult framework is built on two main types of modules: individual branch modules - $g^i(.)$ $(i=\overline{0,M})$, and reconstruction modules - $r^i(.)$ $(i=1,\dots,M)$. 
For the projectors and the fusion module, denoted as $f^i(.)$ $(i=\overline{0,M})$, we adopt designs from previous studies. 
% This approach allows Robult to work with various data types and integrate any appropriate fusion technique. 
In this section, we analyze the complexity of our two proposed modules, which operate in latent spaces and have straightforward designs.
\subsubsection{Individual branches}
Since all $g^i(.)$'s are working with the same input latent space (all raw modalities are projected to the same space), we unify the design of $g^i(.)$'s to be identical across different modalities. Specifically, $g^i(.)$ are constituted by multiple Fully Connected layers, with middle $ReLU$ activations; the last layer of $g^i(.)$ involve no activation, but a L2 normalization operation.
Below are the table of hyper-parameters involved in the analysis.
\begin{table}[ht]
\centering
\begin{tabular}{@{}ll@{}}
\toprule
Notation & Description                             \\ \midrule
$M$        & number of modalities                    \\
$L$        & number of FC layers                     \\
$d_i$      & hidden dimension of $i^th$ layer's output \\
$d_0$      & input dimension                         \\ \bottomrule
\end{tabular}
\caption{$g^i(.)$ related hyper-parameters}
\label{tab:my_label}
\end{table}

\textbf{Time Complexity.} Assume a single operation can be performed in unit time ($\mathcal{O}(1)$). We have the calculation for number of operations in a forward pass as follows.
$$
\begin{aligned}
\text{Within the $i^th$ FC layer:} \\
&d_{i-1}*d_i + di, \\
\text{Over $L$ layers:} \\
&\sum_{i=1}^L d_{i-1}*d_i + di. \\
\end{aligned}
$$
In our implementations, we choose the same dimensions for all hidden outputs (same $d=d_i \forall i=\overline{1,L}$), and there are $M+1$ modules $g^i(.)$. With this, the total number of operation is:
$$
(M+1)\sum_{i=1}^L d_{i-1}*d_i + di = (M+1)*L*d*(d + 1)
=\mathcal{O}(M*L*d^2)
$$
By utilizing matrix product and GPU acceleration, $d^2$ operations can in fact be performed in $\mathcal{O}(1)$ time, make the whole time complexity for individual branches be $\mathcal{O}(M*L)$, which is linearly scaled with $M$.

\textbf{Space Complexity.} Regarding the space complexity, within $i^{th}$ layer, beside the need for storing parameter matrix of size $(d_{i-1} + 1)\times d_i$, output after performing $ReLU$ activation are also stored to later perform back-propagation. Hence, the total number of stored parameters is:
$$(d_{i-1} + 1)*d_i + d_i = (d_{i-1} + 2)*d_i.$$
Following similar derivation with $L$ layers and $M+1$ branches, replacing $d=d_i \forall i=\overline{1,L}$, we have the total space complexity is:
$$(M+1)*L*(d + 2)*d = \mathcal{O}(M*L*d^2).$$

\subsubsection{Reconstruction modules}
 For these reconstruction modules $r^i(.)$'s, we also adopt a similar design patterns as that of individual branches $g^i(.)$. The only differences are the dimension of input for first FC layer ($2d$), which corresponding to the concatenation of $g^0(.)$'s and $g^i(.)$'s outputs. 
 
 \textbf{Time complexity.} With that intuition, as we have $M$ branches and $L$ layers, the total number of calculations is:
 $$
 \begin{aligned}
 &M*\left[(2d + 1)*d + (L-1)*(d+1)*d\right] \\
 = &M*\left[d^2 + L*(d+1)*d\right] = \mathcal{O}(M*L*d^2)
 \end{aligned}
 $$
 Reducing $d^2$ operations to $\mathcal{O}(1)$ time complexity, the same result as observed with $g^i(.)$'s are observed - $\mathcal{O}(M*L)$.

 \textbf{Space complexity.} The total number of stored parameters is:
 $$
 \begin{aligned}
 &M*\left[(2d + 2)*d + (L-1)*(d+2)*d\right] \\
 = &M*\left[(d^2 + L*(d+2)*d\right] = \mathcal{O}(M*L*d^2).
 \end{aligned}
 $$

 In conclusion, all the proposed modules of Robult are linearly scaled (both in time and space), with the number of modalities $M$.

\subsubsection{Computational Time Quantitative Result}
\begin{table}[!th]
\centering
\resizebox{0.45\textwidth}{!}{%
\begin{tabular}{@{}lll@{}}
\toprule
Robult       & GMC          & ActionMAE    \\ \midrule
\multicolumn{3}{l}{\textit{CMU-MOSI:}}              \\
1.08 GFLOPS  & 1.05 GFLOPS  & 1.13 GFLOPS  \\
507.44 MMACs & 492 MMACs    & 526.36 MMACs \\
1.46 M       & 1.16 M       & 11.55 M      \\ \midrule
\multicolumn{3}{l}{\textit{CMU-MOSEI:}}             \\
1.09 GFLOPS  & 1.06 GFLOPS  & 1.15 GFLOPS  \\
508.7 MMACs  & 493.26 MMACs & 526.36 MMACs \\
1.41 M       & 1.17 M       & 11.56 M      \\ \midrule
\multicolumn{3}{l}{\textit{Hateful Memes:}}         \\
32.81 MFLOPs & 26.56 MFLOPS & 61.37 MFLOPS \\
16.39 MMACs  & 13.27 MMACs  & 30.48 MMACs  \\
888.32 K     & 674.05 K     & 1.04 M       \\ \bottomrule
\end{tabular}
}
\caption{Computational times of different methods on different datasets.}
\label{tab:computational}
\end{table}
To further substantiate our results, we measured FLOPs and MACs for several datasets we utilized, comparing them with our current baselines (Table \ref{tab:computational}). For a fair comparison, we kept all unimodal projectors the same. The results suggest that our method introduces slight overheads compared to GMC, but remains faster than ActionMAE, while significantly outperforming both methods in downstream performance.

\section{Implementation details}
\subsection{Environment Settings}
\label{sec:app_compute}
All implementations and experiments are conducted on a single machine equipped with the following hardware configuration: a 6-core Intel Xeon CPU paired with 2 NVIDIA A100 GPUs for accelerated training.

Our codebase predominantly utilizes the \emph{PyTorch 2.0} framework, including the \emph{Pytorch-AutoGrad}, for deep learning model design and calculations. Additionally, we leverage utilities from \emph{Scikit-learn, Pandas,} and \emph{Matplotlib} to support various functionalities in our experiments.
The original codebase for Robult will be made publicly available upon publication.

\subsection{Reproduction and Adaptation}
\label{sec:app_implement}
In both Robult and the Unimodal baselines, we modify the architecture of the projectors $f^{i}(.)$ $(i=\overline{0,M})$ while keeping $g^{i}(.)$ $(i=\overline{0,M})$ as simple as possible. For all testing datasets, the unimodal branches $g^{i}(.)$ consist of a simple Fully Connected layer followed by $L_2$ normalization. In contrast, $g^{0}(.)$ has a higher representation capacity with two Fully Connected layers and ReLU activation.
In the case of GMC \cite{icml22} and ActionMAE \cite{aaai23}, the same architecture of the projectors is adopted as Robult to ensure a fair comparison, with the remaining designs being directly inherited from the original codebases. For Prompt-Trans \cite{cvpr23_2}, we keep all the architecture designs intact and only change the datasets' settings to semi-supervised and missing modalities scenarios. Additional information about the baselines should be best referenced from their original works.

\textbf{CMU-MOSI and CMU-MOSEI datasets.} We follow the settings in \cite{icml22}, which involve temporally-aligned versions of these datasets generated with \cite{mosi_mosei_align}, with additional adaptions for semi-supervised and missing modalities scenarios. Specifically, multimodal Transformer \cite{backbone} is adopt as the joint-modality encoder $f^{0}(.)$ for our model and all state-of-the-art baselines; single-layer GRUs are adopted as unimodal projectors $f^{i}(.)$. The latent space dimension is set as $60$, and all methods are trained in $40$ epochs with Adam optimizer \cite{adam} at the learning rate of $10^{-3}$.

\textbf{MM-IMDb, UPMC Food-101 and Hateful Memes datasets.} 
 For the classification tasks associated with the MM-IMDb, UPMC Food-101, and Hateful Memes datasets, we initially generate text and visual embeddings offline using a pretrained ViLT framework \cite{vilt}. Subsequently, all models are trained using these embeddings instead of raw data. This specific procedure is intentionally conducted to ensure fair evaluation, as Prompt-Trans \cite{cvpr23_2} functioning also involves the same frozen ViLT framework in their training process. We also follow Prompt-Trans \cite{cvpr23_2} for the preprocessing procedures of raw texts and images. For all methods, the offline embedding space's dimension is fixed at $784$ and further condensed into a $128$-dimensional hidden latent space. Additionally, with this setting, we choose the projectors $f^{i}(.)$ $(i=\overline{0,M})$ as simple Fully Connected layers.

\section{Additional empirical results and analysis}
\label{sec:app_exp}
In this sections, we present additional empirical results and analysis to study the behavior of Robult and baselines in different extended settings.

\subsection{Extended modalities missing scenarios}
\label{sec:app_missing}
\begin{table*}[!th]
\centering
% \scriptsize
\setlength\tabcolsep{3pt}
\resizebox{0.65\textwidth}{!}{%
\begin{tabular}{@{}lcccc|cccc@{}}
        & \multicolumn{4}{c}{CMU-MOSI}                                      & \multicolumn{4}{c}{CMU-MOSEI}                                                                                                 \\ \midrule
Metrics & Unimodal & GMC    & ActionMAE & Robult                                & Unimodal                              & GMC                                   & ActionMAE & Robult                                \\ \midrule
\multicolumn{9}{l}{\textit{Text Modality:}}                                                                                                                                                                                                                                      \\
MAE     & 1.41                         & \textcolor{blue}{1.407} & 1.476                        & \textcolor{red}{\textbf{1.397}} & \textcolor{blue}{0.81}                                  & 0.815                                 & 1.115     & \textcolor{red}{\textbf{0.784}} \\
Corr    & 0.137                        & \textcolor{blue}{0.14}  & 0.066                        & \textcolor{red}{\textbf{0.144}} & \textcolor{blue}{0.383}                                 & 0.346                                 & 0.136     & \textcolor{red}{\textbf{0.459}} \\
F1      & 0.551                        & \textcolor{blue}{0.559} & 0.535                        & \textcolor{red}{\textbf{0.578}} & \textcolor{blue}{0.717}                                 & 0.716                                 & 0.614     & \textcolor{red}{\textbf{0.739}} \\
Acc     & 0.553                        & \textcolor{blue}{0.562} & 0.47                         & \textcolor{red}{\textbf{0.569}} & \textcolor{blue}{0.712}                                 & 0.708                                 & 0.603     & \textcolor{red}{\textbf{0.732}} \\ \midrule
\multicolumn{9}{l}{\textit{Audio Modality:}}                                                                                                                                                                                                                                     \\
MAE     & 1.576                        & \textcolor{blue}{1.518} & 1.546                        & \textcolor{red}{\textbf{1.415}} & 0.842                                 & \textcolor{blue}{0.836}          & 1.215     & \textcolor{red}{\textbf{0.825}} \\
Corr    & 0.041                        & -0.065                       & \textcolor{blue}{0.046} & \textcolor{red}{\textbf{0.085}} & 0.111                                 & \textcolor{blue}{0.193}          & 0.101     & \textcolor{red}{\textbf{0.221}} \\
F1      & \textcolor{blue}{0.512} & 0.457                        & 0.508                        & \textcolor{red}{\textbf{0.539}} & 0.618                                 & \textcolor{blue}{0.642}          & 0.634     & \textcolor{red}{\textbf{0.679}} \\
Acc     & \textcolor{blue}{0.496} & 0.46                         & 0.467                        & \textcolor{red}{\textbf{0.535}} & 0.599                                 & \textcolor{blue}{0.63}           & 0.543     & \textcolor{red}{\textbf{0.65}}  \\ \midrule
\multicolumn{9}{l}{\textit{Vision Modality:}}                                                                                                                                                                                                                                    \\
MAE     & \textcolor{blue}{1.451} & 1.497                        & 1.511                        & \textcolor{red}{\textbf{1.425}} & 0.891                                 & \textcolor{blue}{0.839}          & 1.127     & \textcolor{red}{\textbf{0.826}} \\
Corr    & \textcolor{blue}{0.044} & -0.07                        & -0.03                        & \textcolor{red}{\textbf{0.086}} & 0.163                                 & \textcolor{blue}{0.2}            & 0.104     & \textcolor{red}{\textbf{0.201}} \\
F1      & \textcolor{blue}{0.585} & 0.446                        & 0.511                        & \textcolor{red}{\textbf{0.593}} & \textcolor{blue}{0.637}          & 0.621                                 & 0.594     & \textcolor{red}{\textbf{0.647}} \\
Acc     & 0.425                        & 0.449                        & \textcolor{blue}{0.514} & \textcolor{red}{\textbf{0.522}} & \textcolor{blue}{0.624}          & 0.62                                  & 0.561     & \textcolor{red}{\textbf{0.632}} \\ \midrule
\multicolumn{9}{l}{\textit{Text+Audio Modalities:}}                                                                                                                                                                                                                              \\
MAE     & 1.485                        & \textcolor{blue}{1.442} & 1.521                        & \textcolor{red}{\textbf{1.401}} & \textcolor{blue}{0.765}          & 0.813                                 & 1.007     & \textcolor{red}{\textbf{0.762}} \\
Corr    & \textcolor{blue}{0.131} & 0.05                         & 0.089                        & \textcolor{red}{\textbf{0.141}} & \textcolor{blue}{0.418}          & 0.352                                 & 0.202     & \textcolor{red}{\textbf{0.439}} \\
F1      & \textcolor{blue}{0.528} & 0.491                        & 0.507                        & \textcolor{red}{\textbf{0.563}} & \textcolor{blue}{0.729}          & 0.728                                 & 0.624     & \textcolor{red}{\textbf{0.733}} \\
Acc     & \textcolor{blue}{0.512} & 0.493                        & 0.508                        & \textcolor{red}{\textbf{0.546}} & \textcolor{blue}{0.713}          & 0.671                                 & 0.62      & \textcolor{red}{\textbf{0.717}} \\ \midrule
\multicolumn{9}{l}{\textit{Text+Vision Modalities:}}                                                                                                                                                                                                                             \\
MAE     & 1.486                        & \textcolor{blue}{1.465} & 1.489                        & \textcolor{red}{\textbf{1.415}} & 0.861                                 & \textcolor{blue}{0.822}          & 1.003     & \textcolor{red}{\textbf{0.788}} \\
Corr    & \textcolor{blue}{0.144} & 0.044                        & 0.086                        & \textcolor{red}{\textbf{0.146}} & \textcolor{blue}{0.325}          & \textcolor{blue}{0.325}          & 0.198     & \textcolor{red}{\textbf{0.399}} \\
F1      & \textcolor{blue}{0.514} & 0.487                        & 0.501                        & \textcolor{red}{\textbf{0.58}}  & \textcolor{blue}{0.718}          & \textcolor{red}{\textbf{0.722}} & 0.623     & \textcolor{blue}{0.718}          \\
Acc     & 0.492                        & 0.491                        & \textcolor{blue}{0.506} & \textcolor{red}{\textbf{0.534}} & \textcolor{blue}{0.688}          & 0.687                                 & 0.619     & \textcolor{red}{\textbf{0.704}} \\ \midrule
\multicolumn{9}{l}{\textit{Audio+Vision Modalities:}}                                                                                                                                                                                                                            \\
MAE     & \textcolor{blue}{1.432} & 1.499                        & 1.534                        & \textcolor{red}{\textbf{1.426}} & \textcolor{blue}{0.824}          & \textcolor{blue}{0.824}          & 1.173     & \textcolor{red}{\textbf{0.812}} \\
Corr    & \textcolor{blue}{0.014} & -0.075                       & -0.035                       & \textcolor{red}{\textbf{0.091}} & 0.214                                 & \textcolor{blue}{0.233}          & 0.147     & \textcolor{red}{\textbf{0.244}} \\
F1      & 0.492                        & 0.445                        & \textcolor{blue}{0.55}  & \textcolor{red}{\textbf{0.581}} & \textcolor{red}{\textbf{0.748}} & \textcolor{blue}{0.663}          & 0.637     & \textcolor{blue}{0.663}          \\
Acc     & \textcolor{blue}{0.486} & 0.448                        & 0.453                        & \textcolor{red}{\textbf{0.527}} & \textcolor{blue}{0.638}          & 0.633                                 & 0.623     & \textcolor{red}{\textbf{0.64}}  \\ \midrule
\multicolumn{9}{l}{Full Modalities:}                                                                                                                                                                                                                                             \\
MAE     & \textcolor{blue}{1.394} & 1.47                         & 1.496                        & \textcolor{red}{\textbf{1.392}} & \textcolor{blue}{0.783}          & 0.819                                 & 1.103     & \textcolor{red}{\textbf{0.779}} \\
Corr    & \textcolor{blue}{0.186} & 0.101                        & -0.092                       & \textcolor{red}{\textbf{0.247}} & \textcolor{blue}{0.364}          & 0.328                                 & 0.337     & \textcolor{red}{\textbf{0.504}} \\
F1      & \textcolor{blue}{0.597} & 0.497                        & 0.553                        & \textcolor{red}{\textbf{0.657}} & \textcolor{blue}{0.73}           & 0.693                                 & 0.694     & \textcolor{red}{\textbf{0.744}} \\
Acc     & \textcolor{blue}{0.594} & 0.498                        & 0.477                        & \textcolor{red}{\textbf{0.63}}  & \textcolor{blue}{0.729}          & 0.688                                 & 0.643     & \textcolor{red}{\textbf{0.741}} \\ \bottomrule
\end{tabular}
}
\caption{Full performance of different frameworks on CMU-MOSI and CMU-MOSEI Dataset.}
\label{tab:full_mosi_mosei}
\end{table*}
In Table \ref{tab:full_mosi_mosei}, we present the comprehensive performance of different frameworks when provided access to all combinations of input modalities on CMU-MOSI and CMU-MOSEI datasets. This table extends the information presented in Table \ref{tab:mosi_mosei} in the main text. In this experiment, to report the performances of Unimodal baselines and Robult when provided with two modalities, we simply take the mean of the outputs generated by providing these frameworks with single modalities. It is important to note that we do not draw conclusions on the best strategy for merging unimodal results. Despite this, using this simple strategy, Robult consistently produces the best results in most scenarios, highlighting its performance consistency across different missing modality scenarios.

\begin{figure*}[!th]
    \begin{subfigure}[th]{\textwidth}
    \centering
    \includegraphics[width=\textwidth]{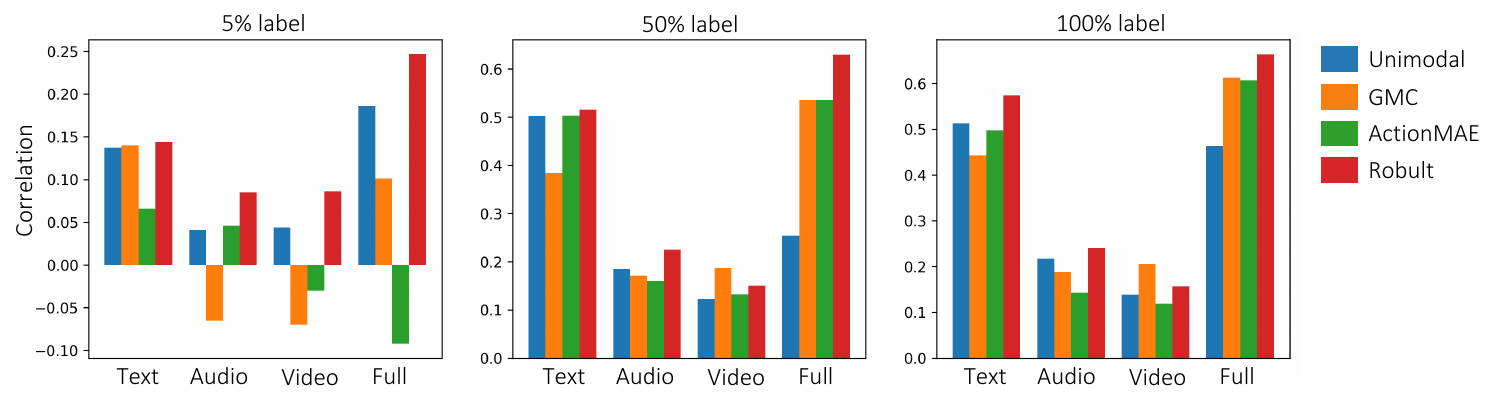}
    \caption{CMU-MOSI dataset}
\end{subfigure}
\begin{subfigure}[th]{\textwidth}
    \centering
    \includegraphics[width=\textwidth]{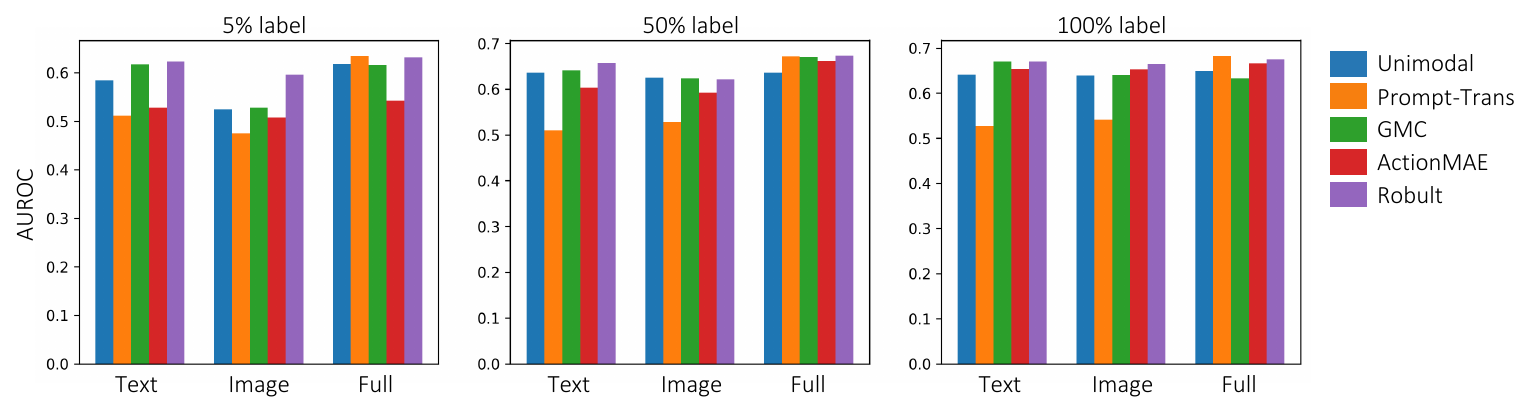}
    \caption{Hateful Memes dataset}
\end{subfigure}
\caption{Models' performances when being exposed to different label ratios.}
\label{fig:semi}
\end{figure*}
\begin{table}[!ht]
\begin{minipage}{\linewidth}
\centering
\resizebox{\linewidth}{!}{%
\begin{tabular}{@{}lllccccc@{}}
\cmidrule(l){2-8}
                         & Modality & Metrics  & \multicolumn{5}{c}{Framework}                                                                                                                                                                \\ \cmidrule(l){2-8}
                                &          &          & Unimodal                              & Prompt-Trans                 & GMC                                   & ActionMAE                             & \textbf{Robult}                       \\ \cmidrule(l){2-8}
                                & Text     & MAE      & \textcolor{blue}{1.157}          & -                            & 1.286                                 & 1.161                                 & \textcolor{red}{\textbf{1.137}} \\
                                &          & Corr     & 0.502                                 & -                            & 0.384                                 & \textcolor{blue}{0.503}          & \textcolor{red}{\textbf{0.516}} \\
                                &          & F1       & \textcolor{blue}{0.712}          & -                            & 0.642                                 & 0.701                                 & \textcolor{red}{\textbf{0.721}} \\
                                &          & Acc      & \textcolor{blue}{0.712}          & -                            & 0.644                                 & 0.709                                 & \textcolor{red}{\textbf{0.722}} \\ \cmidrule(l){2-8} 
                                & Audio    & MAE      & 1.443                                 & -                            & \textcolor{blue}{1.44}           & 1.603                                 & \textcolor{red}{\textbf{1.363}} \\
                                &          & Corr     & \textcolor{blue}{0.185}          & -                            & 0.171                                 & 0.16                                  & \textcolor{red}{\textbf{0.225}} \\
                                &          & F1       & 0.563                                 & -                            & 0.556                                 & \textcolor{blue}{0.566}          & \textcolor{red}{\textbf{0.569}} \\
                                &          & Acc      & \textcolor{blue}{0.559}          & -                            & 0.548                                 & 0.521                                 & \textcolor{red}{\textbf{0.571}} \\ \cmidrule(l){2-8} 
                                & Video    & MAE      & 1.514                                 & -                            & 1.458                                 & \textcolor{red}{\textbf{1.406}} & \textcolor{blue}{1.429}          \\
                                &          & Corr     & 0.123                                 & -                            & \textcolor{red}{\textbf{0.187}} & 0.132                                 & \textcolor{blue}{0.15}           \\
                                &          & F1       & 0.541                                 & -                            & \textcolor{blue}{0.566}          & 0.551                                 & \textcolor{red}{\textbf{0.57}}  \\
                                &          & Acc      & 0.49                                  & -                            & \textcolor{blue}{0.56}           & 0.535                                 & \textcolor{red}{\textbf{0.561}} \\ \cmidrule(l){2-8} 
                                & Full     & MAE      & 1.508                                 & -                            & 1.148                                 & \textcolor{blue}{1.096}          & \textcolor{red}{\textbf{1.092}} \\
                                &          & Corr     & 0.254                                 & -                            & \textcolor{blue}{0.536}          & 0.536                                 & \textcolor{red}{\textbf{0.63}}  \\
                                &          & F1       & 0.566                                 & -                            & 0.709                                 & \textcolor{blue}{0.728}          & \textcolor{red}{\textbf{0.761}} \\
\multirow{-16}{*}{\rotatebox[origin=c]{90}{CMU-MOSI}}     &          & Acc      & 0.545                                 & -                            & \textcolor{blue}{0.71}           & 0.705                                 & \textcolor{red}{\textbf{0.762}} \\ \cmidrule(l){2-8} 
                                & Text     & AUROC    & 0.636                                 & 0.51                         & \textcolor{blue}{0.641}          & 0.603                                 & \textcolor{red}{\textbf{0.657}} \\
                                &          & Accuracy & 0.583                                 & 0.508                        & \textcolor{red}{\textbf{0.592}} & 0.556                                 & \textcolor{blue}{0.589}          \\ \cmidrule(l){2-8} 
                                & Image    & AUROC    & \textcolor{red}{\textbf{0.625}} & 0.528                        & \textcolor{blue}{0.624}          & 0.592                                 & 0.621          \\
                                &          & Accuracy & 0.566                                 & 0.526                        & \textcolor{red}{\textbf{0.568}} & 0.53                                  & \textcolor{red}{\textbf{0.568}} \\ \cmidrule(l){2-8} 
                                & Full     & AUROC    & 0.636                                 & \textcolor{blue}{0.672} & 0.67                                  & 0.661                                 & \textcolor{red}{\textbf{0.673}} \\
\multirow{-6}{*}{\rotatebox[origin=c]{90}{Hateful Memes}} &          & Accuracy & 0.57                                  & 0.594                        & \textcolor{blue}{0.596}          & 0.573                                 & \textcolor{red}{\textbf{0.604}} \\ \cmidrule(l){2-8} 
\end{tabular}
}
\caption{Semi-supervised learning with $50\%$ labelled data on CMU-MOSI and Hateful Memes datasets.}
\label{tab:semi_50}
\end{minipage}
\vfill
\begin{minipage}{\linewidth}
\centering
\resizebox{\linewidth}{!}{%
\begin{tabular}{@{}lllccccc@{}}
\cmidrule(l){2-8}
                                & Modality & Metrics  & \multicolumn{5}{c}{Framework}                                                                                                                                                                         \\ \cmidrule(l){2-8} 
                                &          &          & Unimodal                              & Prompt-Trans                          & GMC                                   & ActionMAE                             & \textbf{Robult}                       \\ \cmidrule(l){2-8} 
                                & Text     & MAE      & {\color[HTML]{0000FF} 1.126}          & -                                     & 1.233                                 & {\color[HTML]{0000FF} 1.108}          & {\color[HTML]{FF0000} \textbf{1.066}} \\
                                &          & Corr     & {\color[HTML]{0000FF} 0.513}          & -                                     & 0.443                                 & {\color[HTML]{0000FF} 0.498}          & {\color[HTML]{FF0000} \textbf{0.574}} \\
                                &          & F1       & {\color[HTML]{0000FF} 0.716}          & -                                     & 0.665                                 & {\color[HTML]{0000FF} 0.739}          & {\color[HTML]{FF0000} \textbf{0.756}} \\
                                &          & Acc      & {\color[HTML]{0000FF} 0.717}          & -                                     & 0.667                                 & {\color[HTML]{0000FF} 0.749}          & {\color[HTML]{FF0000} \textbf{0.753}} \\ \cmidrule(l){2-8} 
                                & Audio    & MAE      & 1.421                                 & -                                     & {\color[HTML]{0000FF} 1.414}          & 1.569                                 & {\color[HTML]{FF0000} \textbf{1.392}} \\
                                &          & Corr     & {\color[HTML]{0000FF} 0.217}          & -                                     & 0.188                                 & 0.143                                 & {\color[HTML]{FF0000} \textbf{0.241}} \\
                                &          & F1       & {\color[HTML]{FF0000} \textbf{0.574}} & -                                     & 0.571                                 & {\color[HTML]{0000FF} 0.569}          & {\color[HTML]{FF0000} \textbf{0.574}} \\
                                &          & Acc      & {\color[HTML]{0000FF} 0.553}          & -                                     & {\color[HTML]{FF0000} \textbf{0.567}} & 0.523                                 & {\color[HTML]{0000FF} \textbf{0.561}} \\ \cmidrule(l){2-8} 
                                & Video    & MAE      & 1.422                                 & -                                     & 1.441                                 & {\color[HTML]{FF0000} \textbf{1.532}} & {\color[HTML]{FF0000} \textbf{1.419}} \\
                                &          & Corr     & 0.138                                 & -                                     & {\color[HTML]{FF0000} \textbf{0.205}} & 0.119                                 & {\color[HTML]{0000FF} 0.156}          \\
                                &          & F1       & {\color[HTML]{0000FF} 0.535}          & -                                     & {\color[HTML]{FF0000} \textbf{0.552}} & 0.534                                 & {\color[HTML]{FF0000} \textbf{0.513}} \\
                                &          & Acc      & {\color[HTML]{0000FF} 0.512}          & -                                     & {\color[HTML]{FF0000} \textbf{0.542}} & 0.421                                 & {\color[HTML]{0000FF} \textbf{0.512}} \\ \cmidrule(l){2-8} 
                                & Full     & MAE      & 1.191                                 & -                                     & 1.093                                 & {\color[HTML]{0000FF} 1.055}          & {\color[HTML]{FF0000} \textbf{1.011}} \\
                                &          & Corr     & 0.463                                 & -                                     & {\color[HTML]{0000FF} 0.612}          & 0.607                                 & {\color[HTML]{FF0000} \textbf{0.663}} \\
                                &          & F1       & 0.726                                 & -                                     & 0.735                                 & {\color[HTML]{0000FF} 0.757}          & {\color[HTML]{FF0000} \textbf{0.764}} \\
\multirow{-16}{*}{\rotatebox[origin=c]{90}{CMU-MOSI}}     &          & Acc      & 0.696                                 & -                                     & {\color[HTML]{0000FF} 0.736}          & 0.763                                 & {\color[HTML]{FF0000} \textbf{0.765}} \\ \cmidrule(l){2-8} 
                                & Text     & AUROC    & 0.641                                 & 0.527                                 & {\color[HTML]{FF0000} \textbf{0.67}}  & 0.654                                 & {\color[HTML]{FF0000} \textbf{0.67}}  \\
                                &          & Accuracy & 0.585                                 & 0.522                                 & {\color[HTML]{0000FF} \textbf{0.607}} & 0.515                                 & {\color[HTML]{FF0000} \textbf{0.63}}  \\ \cmidrule(l){2-8} 
                                & Image    & AUROC    & {\color[HTML]{FF0000} \textbf{0.639}} & 0.541                                 & {\color[HTML]{0000FF} 0.64}           & {\color[HTML]{0000FF} 0.653}          & {\color[HTML]{FF0000} \textbf{0.665}} \\
                                &          & Accuracy & {\color[HTML]{0000FF} 0.595}          & 0.517                                 & {\color[HTML]{FF0000} \textbf{0.591}} & 0.515                                 & {\color[HTML]{FF0000} \textbf{0.622}} \\ \cmidrule(l){2-8} 
                                & Full     & AUROC    & 0.649                                 & {\color[HTML]{FF0000} \textbf{0.683}} & 0.633                                 & 0.666                                 & {\color[HTML]{0000FF} \textbf{0.675}} \\
\multirow{-6}{*}{\rotatebox[origin=c]{90}{Hateful Memes}} &          & Accuracy & 0.585                                 & {\color[HTML]{FF0000} \textbf{0.634}} & {\color[HTML]{0000FF} 0.596}          & 0.536                                 & {\color[HTML]{FF0000} \textbf{0.634}} \\ \cmidrule(l){2-8} 
\end{tabular}
}
\caption{Supervised learning results on CMU-MOSI and Hateful Memes datasets.}
\label{tab:supervised}
\end{minipage}
\end{table}
\subsection{Extended semi-supervised scenarios}
\label{sec:app_semi}

This experiment is designed to observe the behavior of models when exposed to varying amounts of labeled information. In addition to the $5\%$ labeled ratio setting covered in the main text, we additionally evaluate all methods with $50\%$ labeled ratio and ideal supervised settings:
\begin{itemize}
    \item \textbf{Semi-supervised learning with $\mathbf{50\%}$ labelled data.}
Table \ref{tab:semi_50} summarizes the results of the $50\%$ labeled setting with CMU-MOSI and Hateful Memes datasets. As indicated, all methods effectively leverage the increased label signal, resulting in improved performance. However, it's noteworthy that Robult demonstrates its superiority by outperforming other methods on both datasets across various metrics.
\item \textbf{Supervised learning.} With this scenario, we evaluate all methods in an ideal case where fully labelled training dataset is available. Similar to the previous setting, all method further enhance their performance given more labeled data. Robult suggest the consistency by out-performing other baselines in most recorded metrics.
\end{itemize}
To clearly illustrate the performance improvement of Robult and the baselines in each scenario, we provide visualizations of Pearson correlation for CMU-MOSI and AUROC for Hateful Memes in Figure \ref{fig:semi}. Among all methods, Robult demonstrates the best stability and consistency in performance, regardless of input modalities or label ratios.

\subsection{Extended comparison with recent frameworks}
\label{sec:app_sota}
\textbf{Baselines.} We adopt two recent approaches utilizing Constrastive Loss \cite{add_baseline1} and recontruction strategy \cite{add_baseline2} for more comprehensive comparision of Robult with exisiting State-of-the-art frameworks. Their original codebases are slightly adjusted for semi-supervised settings, and the dimensions of the latent space are aligned with Robult's ($60$) to minimize bias in the comparison.
\begin{table}[!th]
\centering
\setlength\tabcolsep{3pt}
\resizebox{\linewidth}{!}{%
\begin{tabular}{@{}llcccccc@{}}
\toprule
                           &                           & \multicolumn{3}{c}{MOSI Dataset}                                                                                                     & \multicolumn{3}{c}{MOSEI Dataset}                                                                            \\ \cmidrule(l){3-5} \cmidrule(l){6-8}
\multirow{-2}{*}{Modality} & \multirow{-2}{*}{Metrics} & \multicolumn{1}{c}{DiCMoR}            & \multicolumn{1}{c}{SEM}      & \multicolumn{1}{c}{Robult}                                    & \multicolumn{1}{c}{DiCMoR}            & \multicolumn{1}{c}{SEM}      & \multicolumn{1}{c}{Robult}            \\ \midrule
Text                       & MAE                       &\textcolor{blue}{ 1.444}          & 1.632                        & \textcolor{red}{\textbf{1.397}}                         &\textcolor{blue}{ 0.819}          & 0.894                        & \textcolor{red}{\textbf{0.784}} \\
                           & Corr                      & 0.085                                 &\textcolor{blue}{ 0.095} & \textcolor{red}{\textbf{0.144}} &\textcolor{blue}{ 0.276}          & 0.252                        & \textcolor{red}{\textbf{0.459}} \\
                           & F1                        &\textcolor{blue}{ 0.536}          & 0.506                        & \textcolor{red}{\textbf{0.578}} & 0.572                                 &\textcolor{blue}{ 0.599} & \textcolor{red}{\textbf{0.739}} \\
                           & Acc                       & 0.511                                 &\textcolor{blue}{ 0.53}  & \textcolor{red}{\textbf{0.569}}                         &\textcolor{blue}{ 0.657}          & 0.63                         & \textcolor{red}{\textbf{0.732}} \\ \midrule
Audio                      & MAE                       &\textcolor{blue}{ 1.504}          & 1.739                        & \textcolor{red}{\textbf{1.415}}                         &\textcolor{blue}{ 0.829}          & 1.037                        & \textcolor{red}{\textbf{0.825}} \\
                           & Corr                      & 0.006                                 &\textcolor{blue}{ 0.043} & \textcolor{red}{\textbf{0.085}}                         &\textcolor{blue}{ 0.201}          & 0.138                        & \textcolor{red}{\textbf{0.221}} \\
                           & F1                        &\textcolor{blue}{ 0.484}          & 0.441                        & \textcolor{red}{\textbf{0.539}}                         & 0.537                                 &\textcolor{blue}{ 0.545} & \textcolor{red}{\textbf{0.679}} \\
                           & Acc                       &\textcolor{blue}{ 0.49}           & 0.463                        & \textcolor{red}{\textbf{0.535}}                         &\textcolor{blue}{ 0.642}          & 0.536                        & \textcolor{red}{\textbf{0.65}}  \\ \midrule
Vision                     & MAE                       &\textcolor{blue}{ 1.454}          & 1.87                         & \textcolor{red}{\textbf{1.425}}                         &\textcolor{blue}{ 0.83}           & 0.992                        & \textcolor{red}{\textbf{0.826}} \\
                           & Corr                      &\textcolor{blue}{ 0.019}          & 0.017                        & \textcolor{red}{\textbf{0.086}}                         &\textcolor{blue}{ 0.163}          & 0.133                        & \textcolor{red}{\textbf{0.201}} \\
                           & F1                        &\textcolor{blue}{ 0.526}          & 0.449                        & \textcolor{red}{\textbf{0.593}}                         &\textcolor{blue}{ 0.552}          & 0.523                        & \textcolor{red}{\textbf{0.647}} \\
                           & Acc                       & \textcolor{red}{\textbf{0.524}} & 0.475                        &\textcolor{blue}{ 0.522}                                  & \textcolor{red}{\textbf{0.635}} & 0.524                        &\textcolor{blue}{ 0.632}          \\ \midrule
Text + Audio               & MAE                       &\textcolor{blue}{ 1.481}          & 1.728                        & \textcolor{red}{\textbf{1.401}} &\textcolor{blue}{ 0.828}          & 0.919                        & \textcolor{red}{\textbf{0.762}} \\
                           & Corr                      & 0.013                                 &\textcolor{blue}{ 0.125} & \textcolor{red}{\textbf{0.141}}                         & 0.173                                 &\textcolor{blue}{ 0.248} & \textcolor{red}{\textbf{0.439}} \\
                           & F1                        &\textcolor{blue}{ 0.494}          & 0.443                        & \textcolor{red}{\textbf{0.563}}                         & 0.563                                 &\textcolor{blue}{ 0.611} & \textcolor{red}{\textbf{0.733}} \\
                           & Acc                       &\textcolor{blue}{ 0.495}          & 0.465                        & \textcolor{red}{\textbf{0.546}}                         &\textcolor{blue}{ 0.639}          & 0.602                        & \textcolor{red}{\textbf{0.717}} \\ \midrule
Text + Vision              & MAE                       &\textcolor{blue}{ 1.473}          & 1.758                        & \textcolor{red}{\textbf{1.415}} &\textcolor{blue}{ 0.832}          & 0.92                         & \textcolor{red}{\textbf{0.788}} \\
                           & Corr                      & 0.012                                 &\textcolor{blue}{ 0.077} & \textcolor{red}{\textbf{0.146}}                         & 0.158                                 &\textcolor{blue}{ 0.25}  & \textcolor{red}{\textbf{0.399}} \\
                           & F1                        &\textcolor{blue}{ 0.514}          & 0.452                        & \textcolor{red}{\textbf{0.58}}                          & 0.579                                 &\textcolor{blue}{ 0.612} & \textcolor{red}{\textbf{0.718}} \\
                           & Acc                       &\textcolor{blue}{ 0.514}          & 0.476                        & \textcolor{red}{\textbf{0.534}}                         &\textcolor{blue}{ 0.638}          & 0.592                        & \textcolor{red}{\textbf{0.704}} \\ \midrule
Audio + Vision             & MAE                       &\textcolor{blue}{ 1.478}          & 1.794                        & \textcolor{red}{\textbf{1.426}} &\textcolor{blue}{ 0.836}          &\textcolor{blue}{ 0.923} & \textcolor{red}{\textbf{0.812}} \\
                           & Corr                      & 0.023                                 &\textcolor{blue}{ 0.035} & \textcolor{red}{\textbf{0.091}}                         & 0.138                                 &\textcolor{blue}{ 0.143} & \textcolor{red}{\textbf{0.244}} \\
                           & F1                        &\textcolor{blue}{ 0.484}          & 0.429                        & \textcolor{red}{\textbf{0.581}}                         &\textcolor{blue}{ 0.581}          & 0.535                        & \textcolor{red}{\textbf{0.663}} \\
                           & Acc                       &\textcolor{blue}{ 0.491}          & 0.451                        & \textcolor{red}{\textbf{0.527}}                         &\textcolor{blue}{ 0.626}          & 0.552                        & \textcolor{red}{\textbf{0.64}}  \\ \midrule
Full                       & MAE                       &\textcolor{blue}{ 1.468}          & 1.797                        & \textcolor{red}{\textbf{1.392}} &\textcolor{blue}{ 0.839}          & 0.902                        & \textcolor{red}{\textbf{0.779}} \\
                           & Corr                      & 0.035                                 &\textcolor{blue}{ 0.041} & \textcolor{red}{\textbf{0.247}}                         & 0.149                                 &\textcolor{blue}{ 0.249} & \textcolor{red}{\textbf{0.504}} \\
                           & F1                        &\textcolor{blue}{ 0.488}          & 0.432                        & \textcolor{red}{\textbf{0.657}}                         & 0.587                                 &\textcolor{blue}{ 0.625} & \textcolor{red}{\textbf{0.744}} \\
                           & Acc                       &\textcolor{blue}{ 0.495}          & 0.453                        & \textcolor{red}{\textbf{0.63}}                          & 0.614                                 &\textcolor{blue}{ 0.667} & \textcolor{red}{\textbf{0.741}} \\ \bottomrule
\end{tabular}
}
\caption{Additional comparison on CMU-MOSI and CMU-MOSEI Datasets.}
\label{tab:add_mosi_mosei}
\end{table}

\textbf{Settings and Result.} We evaluate the models' performance using a $5\%$ semi-supervised task with the CMU-MOSI and CMU-MOSEI datasets, testing all possible combinations of modalities input. Table \ref{tab:add_mosi_mosei} summarizes the results of this study. As shown, Robult consistently outperforms the two frameworks in most scenarios. This experiment further highlights Robult's robustness in semi-supervised settings and when modalities are missing. 

\subsection{Extended ablation studies on Robult design}
\label{sec:app_add_ablation1}
\begin{table}[!th]
\centering
\resizebox{\columnwidth}{!}{%
\begin{tabular}{@{}llrrrr@{}}
\toprule
Modality & Metrics  & \multicolumn{4}{c}{Framework}                                                                                                                                                                                                                                                                                                 \\ \midrule
         &          & \multicolumn{1}{c}{\textbf{Robult}}   & \multicolumn{1}{l}{\begin{tabular}[c]{@{}l@{}}Robult w/o\\ weighting \\ scheme\end{tabular}} & \multicolumn{1}{l}{\begin{tabular}[c]{@{}l@{}}Robult w/o\\ unique\\ branches\end{tabular}} & \multicolumn{1}{l}{\begin{tabular}[c]{@{}l@{}}Robult w/o\\ pseudo\\ labelling\end{tabular}} \\ \midrule
\multicolumn{6}{l}{\textit{MOSI Dataset:}}                                                                                                                                                                                                                                                                                                          \\ 
Text     & MAE      & \color{red}{\textbf{1.397}} & 1.418                                                                                        & 1.514                                                                                    & \color{blue}{ 1.412}                                                                \\
         & Corr     & \color{blue}{0.144}                                 & 0.125                                                                                        & 0.131                                                                                    & \color{red}{\textbf{0.16}}                                                        \\
         & F1       & \color{red}{\textbf{0.578}}                        & 0.551                                                                                        & 0.53                                                                                     & \color{blue}{ 0.553}                                                                \\
         & Acc      & \color{red}{\textbf{0.569}} & 0.548                                                                                        & 0.443                                                                                    & \color{blue}{ 0.551}                                                                \\ \midrule
Audio    & MAE      & \color{red}{\textbf{1.415}} & \color{blue}{ 1.479}                                                                 & 1.576                                                                                    & 1.492                                                                                       \\
         & Corr     & \color{red}{\textbf{0.085}} & -0.042                                                                                       & -0.096                                                                                   & \color{blue}{ -0.001}                                                               \\
         & F1       & \color{red}{\textbf{0.539}} & 0.514                                                                                        & 0.513                                                                                    & \color{blue}{ 0.528}                                                                \\
         & Acc      & \color{red}{\textbf{0.535}} & 0.456                                                                                        & \color{blue}{ 0.514}                                                             & 0.455                                                                                       \\ \midrule
Vision   & MAE      & \color{red}{\textbf{1.425}} & \color{blue}{ 1.434}                                                                 & 1.509                                                                                    & 1.443                                                                                       \\
         & Corr     & \color{blue}{ 0.086}          & \color{red}{\textbf{0.087}}                                                        & 0.034                                                                                    & 0.077                                                                                       \\
         & F1       & \color{red}{\textbf{0.593}} & \color{red}{\textbf{0.593}}                                                        & 0.526                                                                                    & \color{red}{\textbf{0.593}}                                                       \\
         & Acc      & \color{blue}{ 0.522}          & 0.422                                                                                        & \color{red}{\textbf{0.528}}                                                    & 0.422                                                                                       \\ \midrule
Full     & MAE      & 1.392                                 & \color{blue}{ 1.388}                                                                 & 1.487                                                                                    & \color{red}{\textbf{1.359}}                                                       \\
         & Corr     & \color{red}{\textbf{0.247}} & 0.192                                                                                        & 0.207                                                                                    & \color{blue}{ 0.214}                                                                \\
         & F1       & \color{red}{\textbf{0.657}} & 0.566                                                                                        & 0.567                                                                                    & \color{blue}{ 0.595}                                                                \\
         & Acc      & \color{red}{\textbf{0.63}}  & 0.569                                                                                        & 0.496                                                                                    & \color{blue}{ 0.591}                                                                \\ \midrule
\multicolumn{6}{l}{\textit{Hateful Memes:}}                                                                                                                                                                                                                                                                                                         \\
Text     & AUROC    & \color{red}{\textbf{0.623}} & 0.556                                                                                        & \color{blue}{ 0.586}                                                             & 0.555                                                                                       \\
         & Accuracy & \color{red}{\textbf{0.59}}  & 0.541                                                                                        & \color{blue}{ 0.577}                                                             & 0.556                                                                                       \\ \midrule
Image    & AUROC    & 0.596                                 & \color{red}{\textbf{0.597}}                                                        & 0.547                                                                                    & \color{red}{\textbf{0.597}}                                                       \\
         & Accuracy & \color{red}{\textbf{0.562}} & 0.511                                                                                        & \color{blue}{ 0.533}                                                             & 0.51                                                                                        \\ \midrule
Full     & AUROC    & \color{red}{\textbf{0.632}} & 0.571                                                                                        & 0.601                                                                                    & \color{blue}{ 0.602}                                                                \\
         & Accuracy & \color{red}{\textbf{0.595}} & 0.345                                                                                        & \color{blue}{ 0.544}                                                             & 0.51                                                                                        \\ \bottomrule
\end{tabular}
}
\caption{Additional Ablation Study with Robult on two datasets CMU-MOSI and Hateful Memes.}
\label{tab:abl_new_sup}
\end{table}
\textbf{Setting.} In this analysis, our goal is to understand the contributions of our applied strategies to overall Robult's performance. Specifically, we adopt several ablation studies:
\begin{itemize}
    \item \textbf{Removal of Unimodal branches $g^i(.) (i=1\dots M)$}: The output of the shared branch $g^0(.)$ is directly fed into the classifier to yield the final result. The remaining framework is trained normally with the soft PU loss and downstream task loss.
    \item \textbf{Soft-PU Loss Ablation - Removal of weighting scheme}: Uniform weight is adopted instead of our proposed dynamic weighting scheme.
    \item \textbf{Soft-PU Loss Ablation - Removal of pseudo labeling}: All unlabeled samples are considered negatives, resemble normal constrastive learning scheme.
\end{itemize}
\textbf{Result.} The results, presented in Table \ref{tab:abl_new_sup}, indicate an overall performance decrease across all modalities on two tested datasets. 
Specifically, with removal of unimodal branches, in the case of a small dataset with few labeled samples (CMU-MOSI), this ablation causes some weaker modalities to fail in generating beneficial representations during learning. Similar patterns are captures with ablations of Soft P-U loss. The results indicate a consistent decrease in performance across both variations and two test datasets. This analysis empirically supports the effectiveness of our soft PU loss.

\subsection{Extended ablation study regarding choice of RBF Kernel}
\label{sec:app_add_ablation2}
In this analysis, our goal is to understand the role of our weighting scheme in the Soft P-U Loss. We compare two distinct weighting mechanisms to evaluate how closely a positive candidate matches the true positive pair, and then contrast these mechanisms against our initial choice of the RBF Kernel.

\textbf{Setting.} The two new weighting mechanisms are designed based on normalized distances, with the difference lying in the choice of distance measures $\delta(;)$ (here Euclidean and Manhattan distances). Specifically, within a mini-batch $B$, given the reference proximity $\phi_{ref}$ and the proximity $\phi_i$ of the positive candidate that need to be weighted, we calculated the weight as follow:
$$
\begin{aligned}
    w_{i} &= 1 - \Tilde{d}_{i}; \\
    \Tilde{d}_{i} &= \frac{\delta(\phi_i, \phi_{ref})}{\max_{B} \delta(\phi_j, \phi_{ref})}. 
\end{aligned}
$$
\begin{table}[th!]
\centering
\setlength\tabcolsep{3pt}
\resizebox{\columnwidth}{!}{%
\begin{tabular}{@{}llcccccc@{}}
\toprule
                           &                           & \multicolumn{3}{c}{MOSI Dataset}                                                                                                              & \multicolumn{3}{c}{MOSEI Dataset}                                                                                     \\ \cmidrule(l){3-8} 
\multirow{-2}{*}{Modality} & \multirow{-2}{*}{Metrics} & \multicolumn{1}{c}{Robult - L1}       & \multicolumn{1}{c}{Robult - L2}       & \multicolumn{1}{c}{Robult}                                    & \multicolumn{1}{c}{Robult - L1}       & \multicolumn{1}{c}{Robult - L2}       & \multicolumn{1}{c}{Robult}            \\ \midrule
Text                       & MAE                       & 1.486                                 & \textcolor{blue}{ 1.456}          & \textcolor{red}{\textbf{1.397}}                         & 0.793                                 & \textcolor{blue}{ 0.792}          & \textcolor{red}{\textbf{0.784}} \\
                           & Corr                      & 0.1                                   & \textcolor{red}{\textbf{0.184}} & \textcolor{blue}{ 0.144}          & 0.421                                 & \textcolor{blue}{ 0.456}          & \textcolor{red}{\textbf{0.459}} \\
                           & F1                        & 0.571                                 & \textcolor{blue}{ 0.573}          & \textcolor{red}{\textbf{0.578}} & 0.733                                 & \textcolor{red}{\textbf{0.741}} & \textcolor{blue}{ 0.739}          \\
                           & Acc                       & 0.545                                 & \textcolor{red}{\textbf{0.576}} & \textcolor{blue}{ 0.569}                                  & 0.729                                 & \textcolor{red}{\textbf{0.735}} & \textcolor{blue}{ 0.732}          \\ \midrule
Audio                      & MAE                       & \textcolor{blue}{ 1.475}          & 1.51                                  & \textcolor{red}{\textbf{1.415}}                         & \textcolor{red}{\textbf{0.825}} & 0.853                                 & \textcolor{red}{\textbf{0.825}} \\
                           & Corr                      & 0.049                                 & \textcolor{blue}{ 0.083}          & \textcolor{red}{\textbf{0.085}}                         & \textcolor{blue}{ 0.199}          & 0.165                                 & \textcolor{red}{\textbf{0.221}} \\
                           & F1                        & \textcolor{red}{\textbf{0.544}} & 0.477                                 & \textcolor{blue}{ 0.539}                                  & \textcolor{blue}{ 0.674}          & 0.597                                 & \textcolor{red}{\textbf{0.679}} \\
                           & Acc                       & \textcolor{blue}{ 0.52}           & 0.478                                 & \textcolor{red}{\textbf{0.535}}                         & \textcolor{blue}{ 0.635}          & 0.593                                 & \textcolor{red}{\textbf{0.65}}  \\ \midrule
Vision                     & MAE                       & \textcolor{blue}{ 1.475}          & 1.478                                 & \textcolor{red}{\textbf{1.425}}                         & \textcolor{blue}{ 0.917}          & \textcolor{blue}{ 0.931}          & \textcolor{red}{\textbf{0.826}} \\
                           & Corr                      & 0.028                                 & \textcolor{blue}{ 0.045}          & \textcolor{red}{\textbf{0.086}}                         & 0.133                                 & \textcolor{blue}{ 0.18}           & \textcolor{red}{\textbf{0.201}} \\
                           & F1                        & \textcolor{red}{\textbf{0.593}} & 0.582                                 & \textcolor{red}{\textbf{0.593}}                         & 0.567                                 & \textcolor{blue}{ 0.602}          & \textcolor{red}{\textbf{0.647}} \\
                           & Acc                       & 0.492                                 & \textcolor{red}{\textbf{0.522}} & \textcolor{red}{\textbf{0.522}}                         & 0.572                                 & \textcolor{blue}{ 0.603}          & \textcolor{red}{\textbf{0.632}} \\ \midrule
Text + Audio               & MAE                       & 1.477                                 & \textcolor{red}{\textbf{1.395}} & \textcolor{blue}{ 1.401}          & 0.764                                 & \textcolor{red}{\textbf{0.759}} & \textcolor{blue}{ 0.762}          \\
                           & Corr                      & 0.089                                 & \textcolor{red}{\textbf{0.166}} & \textcolor{blue}{ 0.141}                                  & \textcolor{blue}{ 0.439}          & \textcolor{red}{\textbf{0.454}} & \textcolor{blue}{ 0.439}          \\
                           & F1                        & \textcolor{red}{\textbf{0.57}}  & 0.558                                 & \textcolor{blue}{ 0.563}                                  & \textcolor{red}{\textbf{0.74}}  & \textcolor{red}{\textbf{0.74}}  & 0.733                                 \\
                           & Acc                       & 0.531                                 & \textcolor{red}{\textbf{0.561}} & \textcolor{blue}{ 0.546}                                  & \textcolor{blue}{ 0.722}          & \textcolor{red}{\textbf{0.731}} & 0.717                                 \\ \midrule
Text + Vision              & MAE                       & 1.47                                  & \textcolor{red}{\textbf{1.389}} & \textcolor{blue}{ 1.415}          & \textcolor{blue}{ 0.781}          & \textcolor{red}{\textbf{0.772}} & 0.788                                 \\
                           & Corr                      & 0.128                                 & \textcolor{red}{\textbf{0.236}} & \textcolor{blue}{ 0.146}                                  & 0.391                                 & \textcolor{red}{\textbf{0.429}} & \textcolor{blue}{ 0.399}          \\
                           & F1                        & \textcolor{red}{\textbf{0.59}}  & 0.553                                 & \textcolor{blue}{ 0.58}                                   & 0.705                                 & \textcolor{red}{\textbf{0.718}} & \textcolor{red}{\textbf{0.718}} \\
                           & Acc                       & 0.52                                  & \textcolor{red}{\textbf{0.556}} & \textcolor{blue}{ 0.534}                                  & 0.7                                   & \textcolor{red}{\textbf{0.714}} & \textcolor{blue}{ 0.704}          \\ \midrule
Audio + Vision             & MAE                       & \textcolor{blue}{ 1.465}          & 1.475                                 & \textcolor{red}{\textbf{1.426}} & \textcolor{blue}{ 0.834}          & \textcolor{blue}{ 0.843}          & \textcolor{red}{\textbf{0.812}} \\
                           & Corr                      & 0.04                                  & \textcolor{blue}{ 0.054}          & \textcolor{red}{\textbf{0.091}}                         & 0.187                                 & \textcolor{blue}{ 0.216}          & \textcolor{red}{\textbf{0.244}} \\
                           & F1                        & \textcolor{blue}{ 0.577}          & 0.533                                 & \textcolor{red}{\textbf{0.581}}                         & \textcolor{blue}{ 0.635}          & 0.622                                 & \textcolor{red}{\textbf{0.663}} \\
                           & Acc                       & 0.472                                 & \textcolor{blue}{ 0.491}          & \textcolor{red}{\textbf{0.527}}                         & \textcolor{blue}{ 0.622}          & 0.618                                 & \textcolor{red}{\textbf{0.64}}  \\ \midrule
Full                       & MAE                       & 1.403                                 & \textcolor{red}{\textbf{1.366}} & \textcolor{blue}{ 1.392}          & 0.812                                 & \textcolor{red}{\textbf{0.778}} & \textcolor{blue}{ 0.779}          \\
                           & Corr                      & 0.223                                 & \textcolor{blue}{ 0.235}          & \textcolor{red}{\textbf{0.247}}                         & \textcolor{blue}{ 0.45}           & 0.438                                 & \textcolor{red}{\textbf{0.504}} \\
                           & F1                        & 0.554                                 & \textcolor{blue}{ 0.585}          & \textcolor{red}{\textbf{0.657}}                         & 0.703                                 & \textcolor{blue}{ 0.728}          & \textcolor{red}{\textbf{0.744}} \\
                           & Acc                       & 0.547                                 & \textcolor{blue}{ 0.583}          & \textcolor{red}{\textbf{0.63}}                          & 0.708                                 & \textcolor{blue}{ 0.732}          & \textcolor{red}{\textbf{0.741}} \\ \bottomrule
\end{tabular}
}
\caption{Additional ablation study on CMU-MOSI and CMU-MOSEI Datasets.}
\label{tab:add_abl_mosi_mosei}
\end{table}
\textbf{Result.} 
We refer to the variant using a Manhattan distance-based strategy as \textit{Robult - L1}, and the one utilizing Euclidean measures as \textit{Robult - L2}. These two variants are evaluated against the original RBF-based model in a $5\%$ semi-supervised task with the CMU-MOSI and CMU-MOSEI datasets. The comprehensive results are presented in Table \ref{tab:add_abl_mosi_mosei}. 
Generally, we observe minor differences in performance among the weighting schemes. While the RBF approach yields the most consistent results across various input combinations for these datasets, we do not declare it the definitive best weighting method. We believe further research is needed to identify the most appropriate strategy for the dataset of interest.

\subsection{Alignment and Uniformity Analysis}
\label{sec:align_uniform}
\begin{figure*}[ht!]
% \begin{minipage}{0.72\linewidth}
    \centering
    \begin{subfigure}[h]{0.75\textwidth}
    \centering
    \includegraphics[width=\textwidth]{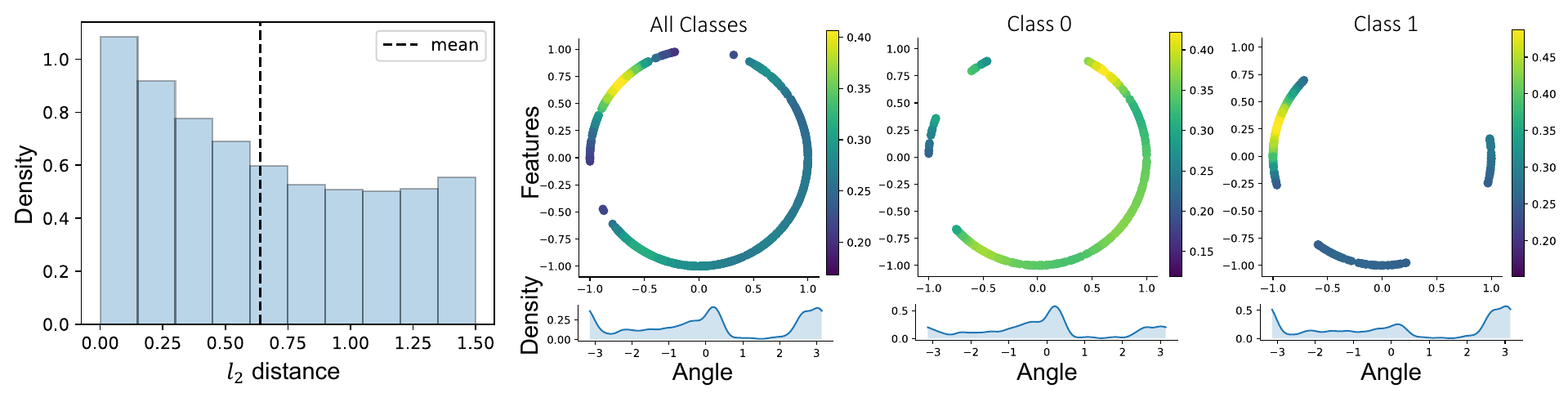}
    \caption{Representations generated with \textbf{Text input}.} 
    \end{subfigure} %
    \begin{subfigure}[h]{0.75\textwidth}
    \centering
    \includegraphics[width=\textwidth]{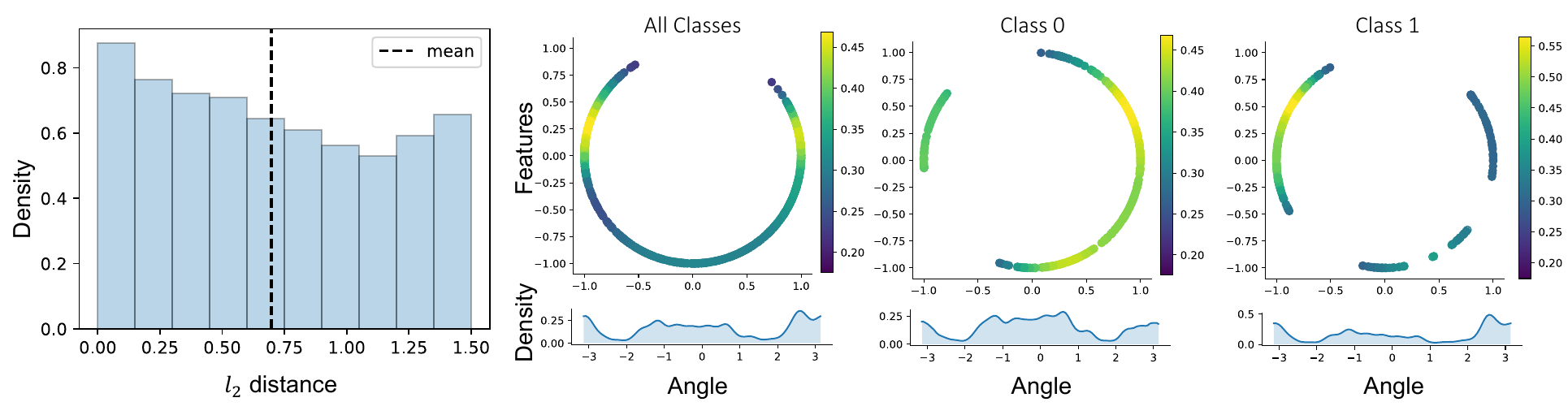}
    \caption{Representations generated with \textbf{Image input}.} 
    \end{subfigure} %
    \begin{subfigure}[h]{0.75\textwidth}
    \centering
    \includegraphics[width=\textwidth]{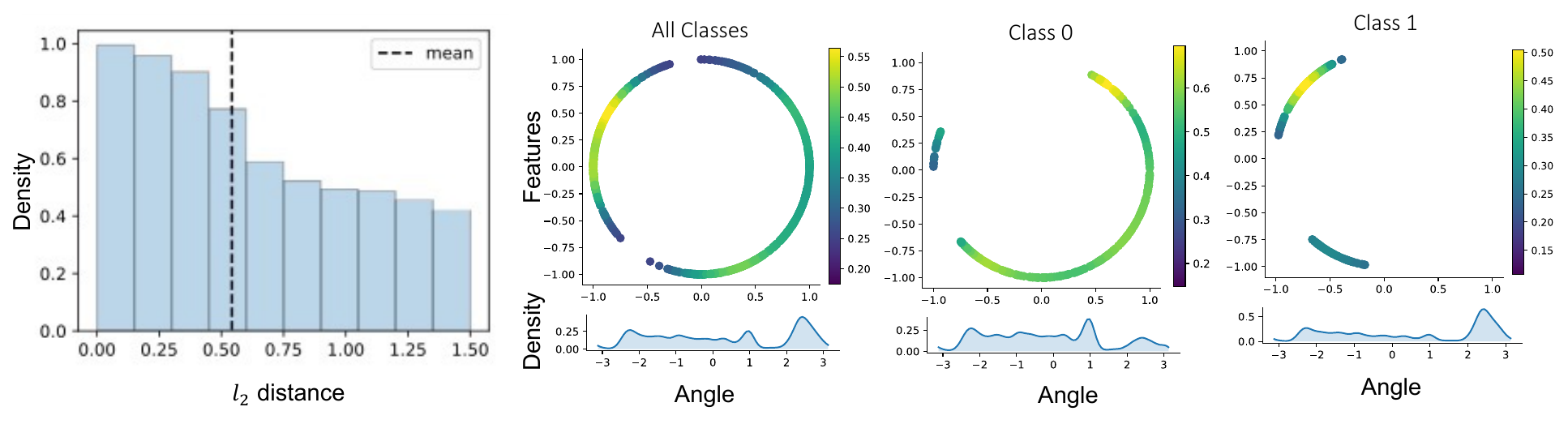}
    \caption{Representations generated with \textbf{full input}.} 
    \end{subfigure}
    \caption{Alignment and Uniformity analysis on representations of Hateful Memes test dataset, generated by Robult.}
    \label{fig:align_unity}
\end{figure*}
We assess the learned representations $Z^i$ and $S$ after the trainining process with soft P-U loss, via two qualities - Alignment and Uniformity \cite{align_uni}.

% Alignment assesses the similarity of features across positive samples, ensuring resistance to noise in those pairs. 
% Conversely, uniformity aims to preserve maximum information by maintaining a uniform feature distribution, simultaneously minimizing the similarities within positive pairs and maximizing the differences between negative pairs. 
Figure \ref{fig:align_unity} provides a comprehensive analysis of the learned representations generated by Robult using both unimodal and multimodal inputs on the Hateful Memes testing set. 
On the left, the Frobenius-norm distance histograms of positive pairs within the test dataset indicate that the representations generated with all the modalities have the smallest mean distances, and as the distances increase, their corresponding density decreases.
% decreasing bin heights as the distance increases.
% The Frobenius-norm distance histograms of positive pairs within the test dataset are visualized, indicating that representations generated with full modalities input have the smallest mean distances and decreasing bin heights as the distance increases. 
While not as compact as the representations with full modalities input, positive pairs' representations generated with unimodal input still exhibit low mean distances and good histogram shapes.
Furthermore, to analyze the uniformity characteristics of the learned representations, we follow the process outlined in \cite{align_uni} and show the result on the right of Figure \ref{fig:align_unity}. The learned representations are projected into $\mathbb{R}^2$ using t-SNE \cite{t_sne}, and the output feature distributions are visualized using Gaussian kernel density estimation (KDE) along with von Mises-Fisher (vMF) KDE for angles ($\texttt{arctan2(y;x)}$). 
As suggested by these figures, Robult's representations demonstrate uniform characteristics on the entire test set as well as good clustering between classes. Specifically, representations of different classes reside on different segments of the unit circle and form separated clusters in Figure \ref{fig:cluster}. The level of separation for different classes with different input modalities correlates well with the actual quantitative results, as shown in Table \ref{tab:cls}. Additional clustering comparison between different method can be found in Appendix \ref{sec:sup_cluster}.

\subsection{Mutual Information Maximization Analysis}
\begin{table}[!th]
\centering
\resizebox{0.6\columnwidth}{!}{%
\begin{tabular}{@{}lcc@{}}
\toprule
Modality             & \multicolumn{2}{l}{\begin{tabular}[c]{@{}l@{}}Mutual Information \\ with fused representation\end{tabular}} \\ \midrule
                     & Robult                 & \begin{tabular}[c]{@{}l@{}}Robult w/o \\ Soft P-U Loss\end{tabular}                \\ \cmidrule(lr){2-3}
Text                 & 0.309                  & 0.054                                                                              \\
Audio                & 0.285                  & 0.077                                                                              \\
Vision               & 0.274                  & 0.083                                                                              \\
Fused & \textbf{2.037}                  & \textbf{1.707}                                                                              \\ \bottomrule
\end{tabular}
}
\caption{Mutual Information between fused and unimodal representations on the CMU-MOSI dataset.}
\label{tab:mi}
\end{table}

As stated in our main text, the necessity to model the objective of learning unimodal representations to maximize mutual information with a lower bound arises because mutual information cannot be precisely calculated. This is due to the changing values of the variables over time and the discrete nature of the datasets. 
To verify the effectiveness of our proposed method, we adopt the histogram-based method in \cite{mi_approx} to approximate MI between two variables after the training process with and without our soft PU loss (Table \ref{tab:mi}). The result suggest two important points:
\begin{itemize}
    \item With our soft PU loss, the mutual information of all unimodal representations with the fused representation increase significantly.
    \item The entropy of the fused representation also increases with the use of our loss, suggesting that the fused representation also get enriched after training with the soft PU loss.
\end{itemize}

\subsection{Soft label quality Analysis}
\label{sec:app_soft_label}
We acknowledge that the quality of pseudo-labels is crucial for effective model training. This is why we incorporate our weighting scheme into the Positive-Unlabeled (PU) contrastive loss, considering the stochastic and unstable nature of pseudo-labels. This approach helps to reduce the impact of noisy pseudo-labels on the training process. 
\begin{figure}[th]
    \centering
    \begin{subfigure}[h]{0.4\textwidth}
    \centering
    \includegraphics[width=\textwidth]{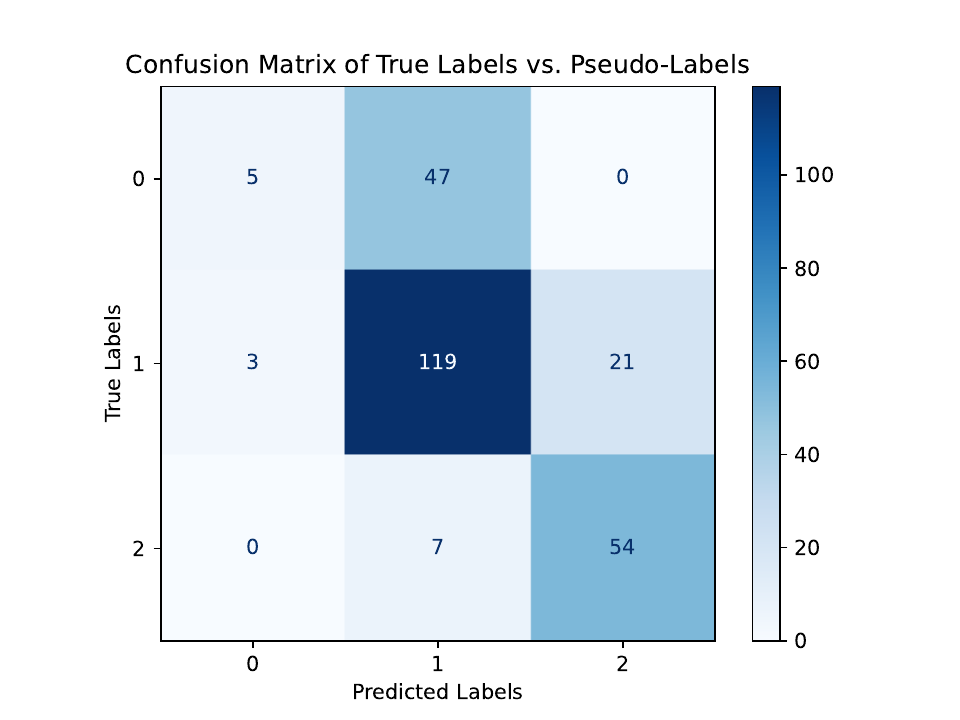}
    \caption{Without weighting scheme} 
    \end{subfigure} %
    \begin{subfigure}[h]{0.4\textwidth}
    \centering
    \includegraphics[width=\textwidth]{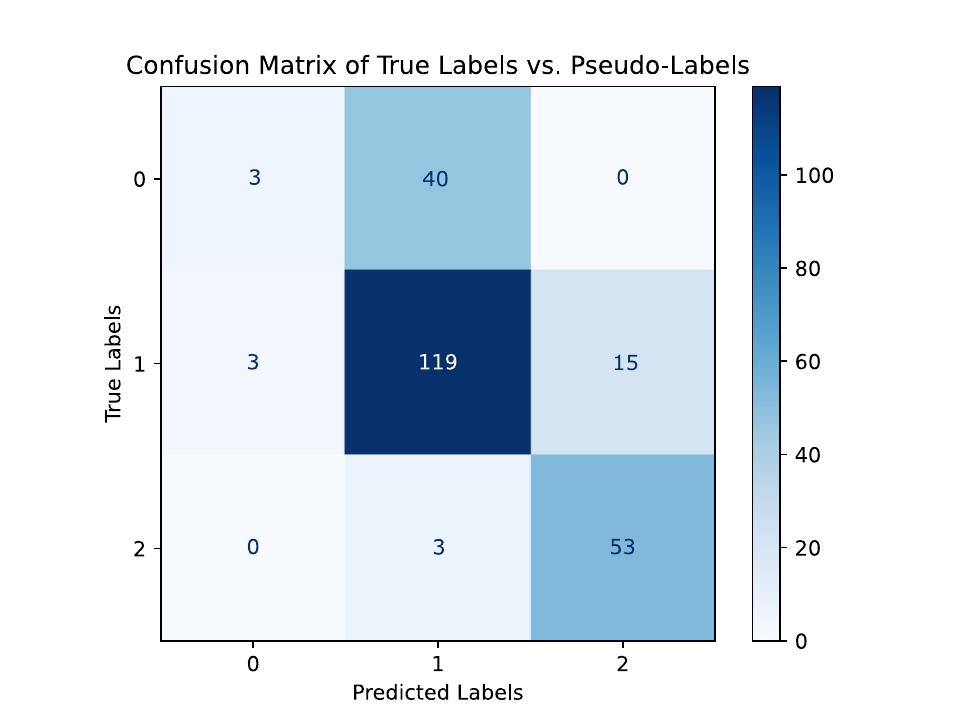}
    \caption{With weighting scheme: Weight threshold of 25$\%$ percentile.} 
    \end{subfigure} %
    \caption{Confusion matrix of Pseudo Labels versus groundtruth label at epoch $20$ on CMU-MOSI dataset.} 
    \label{fig:pseudo_label}
\end{figure}

To demonstrate the effect of both pseudo-labels and our weighting strategy, we visualize the confusion matrix of pseudo-labels with and without the weighting scheme, compared to ground truth labels (Figure \ref{fig:pseudo_label}). This figure is plotted at epoch $20$ of our training process using the CMU-MOSI dataset. The confusion matrix shows a strong correlation between pseudo-labels and ground truth labels, and the weighting scheme (removing all samples with weights below the $25\%$ percentile within the batch) effectively filters out some false positives identified by the pseudo labels.

\subsection{Clusterability Analysis}
\label{sec:sup_cluster}
\begin{figure*}[th]
    \centering
    \begin{subfigure}[h]{0.7\textwidth}
    \centering
    \includegraphics[width=\textwidth]{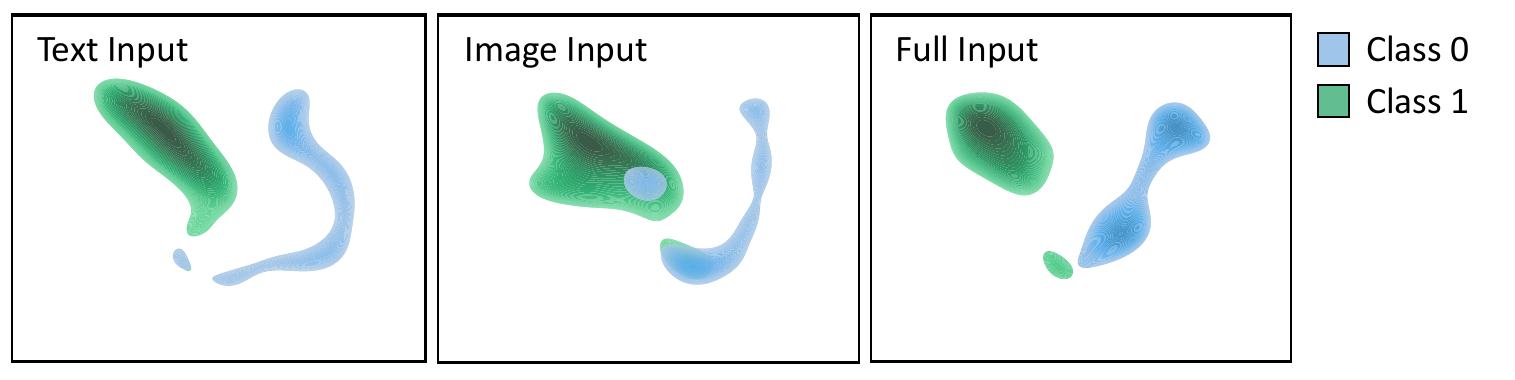}
    \caption{Robult} 
    \end{subfigure} %
    \begin{subfigure}[h]{0.7\textwidth}
    \centering
    \includegraphics[width=\textwidth]{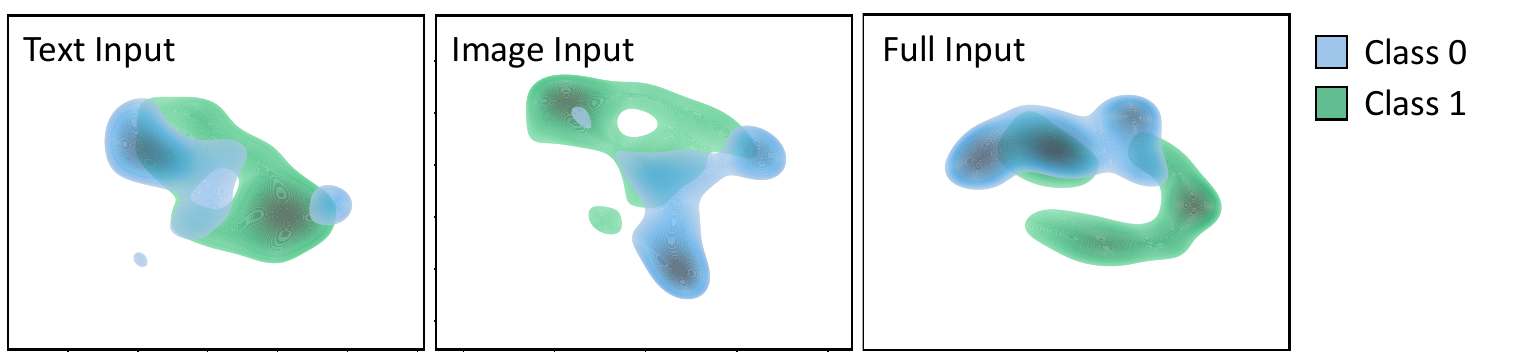}
    \caption{GMC} 
    \end{subfigure} %
    \begin{subfigure}[h]{0.7\textwidth}
    \centering
    \includegraphics[width=\textwidth]{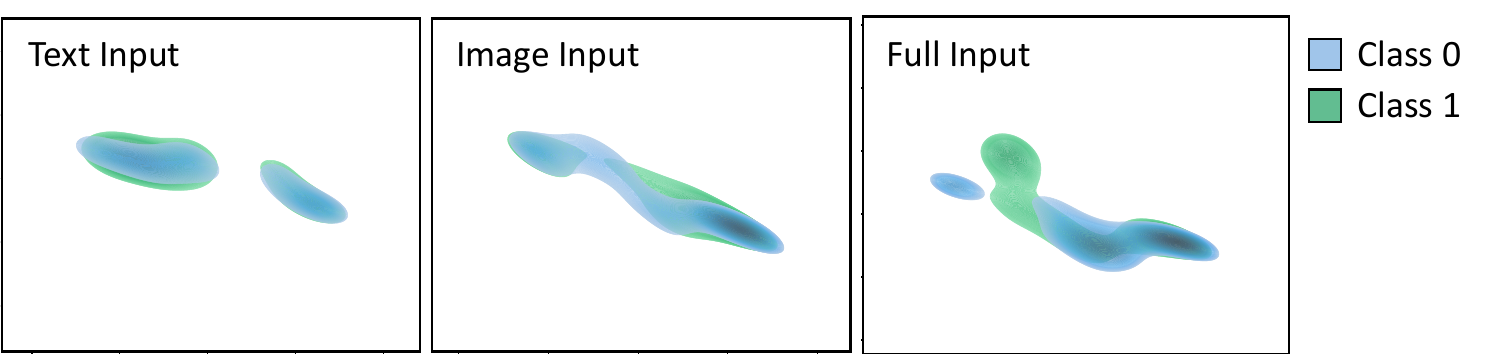}
    \caption{ActionMAE} 
    \end{subfigure}
    \caption{Representation clusters generated by different methods on Hateful Memes dataset.}
    \label{fig:cluster}
\end{figure*}

Complementing the Alignment Uniformity analysis presented in the main manuscript, we provide a comparison of the clusterability characteristic of the learned representations in Figure \ref{fig:cluster}. This experiment is conducted with Robult, compared to GMC and ActionMAE on Hateful Memes dataset. 
This qualitative analysis demonstrates that Robult’s representations are better clustered even in scenarios where different modalities are missing. In contrast, the other methods do not exhibit this level of clustering effectiveness.

\subsection{Transferability Analysis}
\begin{table}[!th]
\centering
% \scriptsize
\setlength\tabcolsep{3pt}
\resizebox{\columnwidth}{!}{%
\begin{tabular}{@{}lcccc@{}}
\toprule
                                  & \multicolumn{4}{c}{MOSI Dataset}                                                                                                                              \\ \midrule
Metrics                           & \multicolumn{1}{c}{Unimodal}          & \multicolumn{1}{c}{GMC}               & \multicolumn{1}{c}{ActionMAE}         & \multicolumn{1}{c}{Robult}            \\ \midrule
\textit{Text Modality:}           & \textcolor{blue}{}               &                                       &                                       & \textcolor{red}{\textbf{}}      \\
MAE                               & \textcolor{blue}{1.448}          & 1.454                                 & 1.456                                 & \textcolor{red}{\textbf{1.446}} \\
Corr                              & \textcolor{red}{\textbf{0.132}} & 0.119                                 & {\color[HTML]{333333} 0.106}          & \textcolor{red}{\textbf{0.132}} \\
F1                                & \textcolor{blue}{0.559}          & 0.551                                 & \textcolor{red}{\textbf{0.624}} & 0.53                                  \\
Acc                               & \textcolor{blue}{0.527}          & 0.48                                  & 0.481                                 & \textcolor{red}{\textbf{0.531}} \\ \midrule
\textit{Audio Modality:}          & \multicolumn{1}{l}{}                  & \multicolumn{1}{l}{}                  & \multicolumn{1}{l}{}                  & \multicolumn{1}{l}{}                  \\
MAE                               & 1.514                                 & \textcolor{red}{\textbf{1.456}} & 1.517                                 & \textcolor{blue}{1.504}          \\
Corr                              & -0.196                                & \textcolor{red}{\textbf{0.085}} & \textcolor{blue}{0.084}          & -0.126                                \\
F1                                & 0.5                                   & \textcolor{red}{\textbf{0.592}} & \textcolor{blue}{0.57}           & 0.486                                 \\
Acc                               & \textcolor{red}{\textbf{0.512}} & 0.429                                 & 0.423                                 & \textcolor{blue}{0.486}          \\ \midrule
\textit{Vision Modality:}         & \multicolumn{1}{l}{}                  & \multicolumn{1}{l}{}                  & \multicolumn{1}{l}{}                  & \multicolumn{1}{l}{}                  \\
MAE                               & \textcolor{blue}{1.473}          & 1.86                                  & 1.547                                 & \textcolor{red}{\textbf{1.38}}  \\
Corr                              & \textcolor{red}{\textbf{0.076}} & -0.087                                & 0.017                                 & \textcolor{blue}{0.058}          \\
F1                                & \textcolor{blue}{0.71}           & 0.593                                 & 0.561                                 & \textcolor{red}{\textbf{0.732}} \\
Acc                               & \textcolor{blue}{0.546}          & 0.422                                 & 0.432                                 & \textcolor{red}{\textbf{0.577}} \\ \midrule
\textit{Text+Audio Modalities:}   & \multicolumn{1}{l}{}                  & \multicolumn{1}{l}{}                  & \multicolumn{1}{l}{}                  & \multicolumn{1}{l}{}                  \\
MAE                               & \textcolor{blue}{1.405}          & 1.441                                 & \textcolor{red}{\textbf{1.378}} & 1.438                                 \\
Corr                              & 0.013                                 & \textcolor{red}{\textbf{0.164}} & \textcolor{blue}{0.101}          & 0.044                                 \\
F1                                & 0.504                                 & \textcolor{blue}{0.569}          & \textcolor{red}{\textbf{0.591}} & 0.516                                 \\
Acc                               & 0.456                                 & 0.467                                 & \textcolor{blue}{0.49}           & \textcolor{red}{\textbf{0.509}} \\ \midrule
\textit{Text+Vision Modalities:}  & \multicolumn{1}{l}{}                  & \multicolumn{1}{l}{}                  & \multicolumn{1}{l}{}                  & \multicolumn{1}{l}{}                  \\
MAE                               & 1.45                                  & 1.629                                 & \textcolor{blue}{1.403}          & \textcolor{red}{\textbf{1.383}} \\
Corr                              & \textcolor{blue}{0.119}          & 0.106                                 & 0.107                                 & \textcolor{red}{\textbf{0.137}} \\
F1                                & \textcolor{red}{\textbf{0.621}} & 0.593                                 & 0.606                                 & \textcolor{blue}{0.615}          \\
Acc                               & 0.536                                 & 0.422                                 & \textcolor{blue}{0.568}          & \textcolor{red}{\textbf{0.575}} \\ \midrule
\textit{Audio+Vision Modalities:} & \multicolumn{1}{l}{}                  & \multicolumn{1}{l}{}                  & \multicolumn{1}{l}{}                  & \multicolumn{1}{l}{}                  \\
MAE                               & \textcolor{blue}{1.434}          & 1.57                                  & 1.528                                 & \textcolor{red}{\textbf{1.428}} \\
Corr                              & -0.181                                & \textcolor{blue}{-0.005}         & \textcolor{red}{\textbf{0.08}}  & -0.088                                \\
F1                                & 0.496                                 & 0.533                                 & \textcolor{blue}{0.551}          & \textcolor{red}{\textbf{0.554}} \\
Acc                               & \textcolor{blue}{0.459}          & 0.422                                 & 0.441                                 & \textcolor{red}{\textbf{0.487}} \\ \midrule
\textit{Full Modalities:}                 & \multicolumn{1}{l}{}                  & \multicolumn{1}{l}{}                  & \multicolumn{1}{l}{}                  & \multicolumn{1}{l}{}                  \\
MAE                               & \textcolor{blue}{1.456}          & 1.684                                 & 1.472                                 & \textcolor{red}{\textbf{1.399}} \\
Corr                              & \textcolor{blue}{0.088}          & 0.072                                 & 0.066                                 & \textcolor{red}{\textbf{0.202}} \\
F1                                & 0.549                                 & \textcolor{blue}{0.573}          & 0.55                                  & \textcolor{red}{\textbf{0.588}} \\
Acc                               & \textcolor{blue}{0.551}          & 0.425                                 & \textcolor{blue}{0.551}          & \textcolor{red}{\textbf{0.585}} \\ \bottomrule
\end{tabular}
}
\caption{Transferability result on CMU-MOSI dataset.}
\label{tab:transfer}
\end{table}
With this experiment, we investigate the tranferability characteristic of Robult, as well as existing state-of-the-art frameworks and baselines. 

\textbf{Experiment settings.} Inspired by common pre-training procedures, where a model is initially trained on a large dataset for a source task and then fine-tuned for a target task, we designed an experiment to evaluate the zero-shot performance of all models on CMU-MOSI after being trained with the CMU-MOSEI dataset. This setting aligns with common practices, as CMU-MOSEI is larger in scale, covering a wider range of sentiment levels and emotions compared to CMU-MOSI \cite{mosei}. 
To conduct the experiment, we first pretrain all methods with CMU-MOSEI using $5\%$ labeled data, simultaneously evaluating them in a semi-supervised scenario. Since zero-shot evaluation requires no fine-tuning stage, all model architectures must remain intact after pretraining. However, there are discrepancies in the dimensions of the input data between CMU-MOSI and CMU-MOSEI. Specifically, the audio input of CMU-MOSI has a latent dimension of $5$, while that of CMU-MOSEI is $74$. Additionally, the video input of CMU-MOSI and CMU-MOSEI is $20$ and $35$, respectively. To address this issue, we generate a compact version of CMU-MOSEI by employing T-SNE on the original data, aligning the dimensions with those in the CMU-MOSI datasets. After pretraining, we directly run evaluations on the normal test set of CMU-MOSI, given different combinations of input modalities to evaluate modalities missing performance.

\textbf{Results.} Table \ref{tab:transfer} provides a summary of the results from this experiment. Generally, all methods experience a reduction in performance in certain cases when transferred to a different dataset. However, among all approaches, Robult consistently achieves the best performance, as indicated by the recorded metrics. In addition, it is noteworthy that Robult is the only approach capable of producing meaningful results with input from full modalities in this zero-shot transfer setting.

\subsection{Incorporation with existing approaches}
This analysis investigates the ability of Robust in incorporating with other approaches to enhance their desired characteristics in learned representations. 

\textbf{Experiment settings.} We select GMC as a baseline approach for conducting this experiment. GMC aims to preserve the geometrical alignment of representations from different modalities through a geometrical contrastive loss \cite{icml22}. To observe the impact of incorporating Robult with GMC to preserve this characteristic, we simply adopt their geometrical contrastive loss with our existing $\mathcal{L}_{(u)lb}$:
\begin{equation}
    \begin{aligned}
        \mathcal{L}_{lb}^i = -\frac{1}{||B_{1, 1}||} \sum_{\substack{(j, k) \sim \\p(F=1, L=1)}} \log v(s_j,z^i_k) &+ \log v(s_j,s_k) \\
        &+ \log v(z^i_j,z^i_k);  \\
         \mathcal{L}_{lb} = -\frac{1}{M} \sum_{i=1}^M \mathcal{L}_{lb}^i. \\
   \end{aligned}
\end{equation}
and: 
\begin{equation}
    \begin{aligned}
         \mathcal{L}_{ulb}^i = -\frac{1}{||B_{1, 0}||} \sum_{\substack{(j,k) \sim \\p(F=1, L=0)}}w_{jk}^i ( \log v(s_j,z^i_k) &+ \log v(s_j,s_k) \\
         &+ \log v(z^i_j,z^i_k)); \\
         \mathcal{L}_{ulb} = -\frac{1}{M} \sum_{i=1}^M \mathcal{L}_{ulb}^i.
    \end{aligned}
\end{equation}
To evaluate the geometrical alignment of the learned representations, we employ Delaunay Component Analysis (DCA) \cite{dca}, a technique similar to that used in GMC. DCA involves comparing geometric and topological properties of an evaluation set of representations (E) with a reference set (R), which acts as an approximation of the true underlying manifold. Following the evaluation strategy outlined in \cite{icml22}, we consider three metrics provided by DCA that reflect the geometric alignment between R (representations of full modalities input) and E (representations of single modality inputs): network quality $q \in [0, 1]$, precision $\mathcal{P}$, and recall $\mathcal{R}$. We report the harmonic mean defined as $3/(1/\mathcal{P} + 1/\mathcal{R} + 1/q)$ when all $\mathcal{P}, \mathcal{R}, q > 0$ and 0 otherwise. For a detailed description of DCA and its settings, please refer to the original work \cite{dca,icml22}. 

\textbf{Results.} We provide the alignment metrics for the representations generated with CMU-MOSI and Hateful Memes datasets, considering only $50\%$ labeled data in their respective training sets (Table \ref{tab:dca}). The statistics indicate that Robult effectively enhances the performance of GMC in its effort to preserve geometrical alignment under the constraint of limited label information. We anticipate that this behavior can potentially be extended to other methods under limited available label information, although additional investigations are needed to verify this.
\begin{table}[!th]
\centering
\resizebox{\columnwidth}{!}{%
\begin{tabular}{@{}llllccc@{}}
\toprule
Dataset       & R    & E      & Metrics & \multicolumn{1}{c}{Unimodal} & \multicolumn{1}{c}{GMC}               & \multicolumn{1}{c}{\textbf{Robult + GMC}}   \\ \midrule
MOSI Dataset  & Full & Text   & MAE     & 0.473                        & \textcolor{blue}{0.529}          & \textcolor{red}{\textbf{0.535}} \\
              & Full & Audio  & Corr    & 0                            & \textcolor{blue}{0.375}          & \textcolor{red}{\textbf{0.393}} \\
              & Full & Vision & F1      & 0                            & \textcolor{red}{\textbf{0.478}} & \textcolor{blue}{0.335}          \\ \midrule
Hateful Memes & Full & Text   & AUROC   & 0.349                            & \textcolor{blue}{0.489}          & \textcolor{red}{\textbf{0.518}} \\
              & Full & Image  & AUROC   & 0                        & \textcolor{blue}{0.456}          & \textcolor{red}{\textbf{0.509}} \\ \bottomrule
\end{tabular}
}
\caption{DCA Scores of models, evaluating geometrical alignment of full-modalities representations with unimodal representations.}
\label{tab:dca}
\end{table}
% \newpage
% \pagenumbering{gobble}
% \input{sections/supplement}
% % \input{sections/additional_supplementary}
% \input{sections/checklist}
\end{document}